\documentclass[10pt,twocolumn,letterpaper]{article}

\usepackage[pagenumbers]{cvpr} % To force page numbers, e.g. for an arXiv version
\usepackage[table]{xcolor}
\usepackage[most]{tcolorbox}
\usepackage{booktabs}
\usepackage{tikz}
\usetikzlibrary{shapes.geometric, arrows.meta, positioning, fit, backgrounds}

\newtcolorbox{disclaimerbox}{
    colback=gray!10,     % background color
    colframe=gray!40,    % frame color
    boxrule=0.5pt,       % border thickness
    arc=4pt,             % rounded corners
    auto outer arc,
    boxsep=5pt,
    left=6pt,
    right=6pt,
    top=4pt,
    bottom=4pt,
    enhanced jigsaw
}

\definecolor{iccvblue}{rgb}{0.21,0.49,0.74}
\usepackage[pagebackref,breaklinks,colorlinks,allcolors=iccvblue]{hyperref}

\def\paperID{*****} % *** Enter the Paper ID here
\def\confName{CVPR}
\def\confYear{2026}

\title{Low Light Image Enhancement Challenge at NTIRE 2026}

\author{
George Ciubotariu$^{\dagger}$ \and
Sharif S M A$^{\dagger}$ \and
Abdur Rehman$^{\dagger}$ \and
Fayaz Ali Dharejo$^{\dagger}$ \and
Rizwan Ali Naqvi$^{\dagger}$ \and
Marcos V.\ Conde$^{\dagger}$ \and
Radu Timofte$^{\dagger}$ \and
Zhi Jin \and
Hongjun Wu \and
Wenjian Zhang \and
Chang Ye \and
Xunpeng Yi \and
Qinglong Yan \and
Yibing Zhang \and
Zaynab Ali \and
Saiprasad Meesiyawar \and
Varda I Pattanshetty \and
Varsha I Pattanshetty \and
Nikhil Akalwadi \and
Padmashree Desai \and
Ramesh Ashok Tabib \and
Uma Mudenagudi \and
Hao Yang \and
Ruikun Zhang \and
Liyuan Pan \and
Furkan Kınlı \and
Donghun Ryou \and
Inju Ha \and
Junoh Kang \and
Bohyung Han \and
Wei Zhou \and
Yuval Haitman \and
Ariel Lapid \and
Reuven Peretz \and
Idit Diamant \and
Leilei Cao \and
Shuo Zhang \and
Praful Hambarde \and
Prateek Shaily \and
Jayant Kumar \and
Hardik Sharma \and
Aashish Negi \and
Sachin Chaudhary \and
Akshay Dudhane \and
Amit Shukla \and
MoHao Wu \and
Lin Wang \and
Jiachen Tu \and
Guoyi Xu \and
Yaoxin Jiang \and
Jiajia Liu \and
Yaokun Shi \and
Raul Balmez \and
Alexandru Brateanu \and
Ciprian Orhei \and
Cosmin Ancuti \and
Codruta O. Ancuti \and
Bilel Benjdira \and
Anas M. Ali \and
Wadii Boulila \and
Kaifan Qiao \and
Bofei Chen \and
Jingyi Xu \and
Duo Zhang \and
Xin Deng \and
Mai Xu \and
Shengxi Li \and
Lai Jiang \and
Harini A \and
Ananya N \and
Lakshanya K \and
Ying Xu \and
Xinyi Zhu \and
Shijun Shi \and
Jiangning Zhang \and
Yong Liu \and
Kai Hu \and
Jing Xu \and
Xianfang Zeng \and
Jinao Song \and
Guangsheng Tang \and
Cheng Li \and
Yuqiang Yang \and
Ziyi Wang \and
Yan Chen \and
Long Bao \and
Heng Sun \and
Mohab Kishawy \and
Jun Chen \and
Wan-Chi Siu \and
Yihao Cheng \and
Hon Man Hammond Lee \and
Chun-Chuen Hui
}

\begin{document}
\maketitle

\begin{abstract}

This paper presents a comprehensive review of the NTIRE 2026 Low Light Image Enhancement Challenge, highlighting the proposed solutions and final results.
The objective of this challenge is to identify effective networks capable of producing clearer and visually compelling images in diverse and challenging conditions by learning representative visual cues with the purpose of restoring information loss due to low-contrast and noisy images. A total of 195 participants registered for the first track and 153 for the second track of the competition, and 22 teams ultimately submitted valid entries. This paper thoroughly evaluates the state-of-the-art advances in (joint denoising and) low-light image enhancement, showcasing the significant progress in the field, while leveraging samples of our novel dataset.
\vspace{-.12cm}
\end{abstract}

{\let\thefootnote\relax\footnotetext{%
\hspace{-5mm} 
$\dagger$
George Ciubotariu, 
Sharif S M A, 
Abdur Rehman, 
Fayaz Ali Dharejo, 
Rizwan Ali Naqvi, 
Marcos V.\ Conde, and
Radu Timofte
 are the NTIRE 
2026
challenge organizers. The other authors participated in the challenge.
\\
Appendix \ref{sec:apd:team} (in the appendix) provides names and affiliations.
% Participants' names and affiliations are provided in the supplementary.
\\ 
NTIRE: \url{https://cvlai.net/ntire/2026}
\\ 
Code: \url{https://github.com/sharif-apu/LSD-TFFormer}}}

\vspace{-.5cm}
\section{Introduction}
\label{sec:intro}

% \textcolor{red}{SLLIE}
Low-light image enhancement (LLIE) remains a fundamental and challenging computer vision task, crucial for applications ranging from consumer photography to autonomous navigation. Despite advancements in computational photography, many standard devices still lack dedicated low-light capabilities \cite{sharif2026illuminating}. Consequently, countless archived images suffer from severe degradation, detail loss, and poor contrast, driving the need for robust algorithms to recover high-quality visual information.

The primary bottleneck in developing robust, generalizable LLIE models stems from the inherent limitations of existing single-shot LLIE (SLLIE) training datasets, such as LOL-V1 \cite{wei2018deep}, LOL-V2 \cite{yang2021sparse}, and LSRW \cite{hai2023r2rnet}. These datasets typically provide a limited number of samples captured in strictly controlled environments, often relying on low ISO settings or artificially dimmed bright-light scenes \cite{sharif2026illuminating}. Consequently, models trained on such constrained data struggle to generalize to complex, uncontrolled real-world environments and frequently fail to enhance the legacy images captured by standard devices.

To bridge this critical gap between synthetic training and real-world application, we present the NTIRE 2026 Low-Light Image Enhancement Challenge. This challenge aims to encourage innovative solutions capable of tackling the complexities of in-the-wild single-shot low-light image enhancement (SLLIE). The competition comprises two distinct tracks built upon the Low-light Smartphone Dataset (LSD) \cite{sharif2026illuminating}. Captured in diverse, uncontrolled environments using 15 smartphone cameras under extreme lighting conditions (0.1–200 lux), LSD provides a robust foundation for learning and evaluating real-world generalization. Track 1 focuses on the visibility and color enhancement of already denoised low-light images, challenging models to recover structural details and accurate colors from standard captures. Track 2 introduces the dual challenge of joint low-light denoising and enhancement; by preserving authentic sensor noise, this track accurately reflects the true complexities of everyday photography. In this report, we detail the challenge design, present the final quantitative and qualitative results, and review the state-of-the-art solutions developed by the participating teams.

% \textcolor{red}{
This challenge is one of the challenges associated with the NTIRE 2026 Workshop on:
deepfake detection~\cite{ntire26deepfake}, 
high-resolution depth~\cite{ntire26hrdepth},
multi-exposure image fusion~\cite{ntire26raim_fusion}, 
AI flash portrait~\cite{ntire26raim_portrait}, 
professional image quality assessment~\cite{ntire26raim_piqa},
light field super-resolution~\cite{ntire26lightsr},
3D content super-resolution~\cite{ntire263dsr},
bitstream-corrupted video restoration~\cite{ntire26videores},
X-AIGC quality assessment~\cite{ntire26XAIGCqa},
shadow removal~\cite{ntire26shadow},
ambient lighting normalization~\cite{ntire26lightnorm},
controllable Bokeh rendering~\cite{ntire26bokeh},
rip current detection and segmentation~\cite{ntire26ripdetseg},
high FPS video frame interpolation~\cite{ntire26highfps},
Night-time dehazing~\cite{ntire26nthaze,ntire26nthaze_rep},
learned ISP with unpaired data~\cite{ntire26isp},
short-form UGC video restoration~\cite{ntire26ugcvideo},
raindrop removal for dual-focused images~\cite{ntire26dual_focus},
image super-resolution (x4)~\cite{ntire26srx4},
photography retouching transfer~\cite{ntire26retouching},
mobile real-word super-resolution~\cite{ntire26rwsr},
remote sensing infrared super-resolution~\cite{ntire26rsirsr},
AI-Generated image detection~\cite{ntire26aigendet},
cross-domain few-shot object detection~\cite{ntire26cdfsod},
financial receipt restoration and reasoning~\cite{ntire26finrec},
real-world face restoration~\cite{ntire26faceres},
reflection removal~\cite{ntire26reflection},
anomaly detection of face enhancement~\cite{ntire26anomalydet},
video saliency prediction~\cite{ntire26videosal},
efficient super-resolution~\cite{ntire26effsr},
3d restoration and reconstruction in adverse conditions~\cite{ntire26realx3d},
image denoising~\cite{ntire26denoising},
blind computational aberration correction~\cite{ntire26aberration},
event-based image deblurring~\cite{ntire26eventblurr},
efficient burst HDR and restoration~\cite{ntire26bursthdr},
low-light enhancement: `twilight cowboy'~\cite{ntire26twilight},
and efficient low light image enhancement~\cite{ntire26effllie}.
% }
\section{Challenge}
\label{sec:challenge}

\paragraph{Dataset.} 
Our challenge utilizes the Low-light Smartphone Dataset (LSD) \cite{sharif2026illuminating} with two input variants: Denoised Low-light Input (DLI) for Track 1 and Noisy Low-light Input (NLI) for Track 2. Training data was generated by extracting $512 \times 512$ patches from 6,025 high-resolution (4K+) scenes. After rigorous reference-guided refinement to remove uninformative or blurry regions, we obtained 115,376 patches for Track 1 and 64,318 for Track 2. For evaluation, we provided 24 high-resolution validation pairs per track. The strictly hidden testing set comprised 26 paired full-resolution images for quantitative evaluation and 60 unpaired real-world images for perceptual assessment per track.

\paragraph{Phases.} 
The challenge consisted of two phases. \textbf{(1) Development and Validation:} Participants received the training data and 24 validation inputs per track (with ground-truth withheld). An evaluation server provided real-time PSNR and SSIM feedback. \textbf{(2) Testing:} Participants received 86 test images per track (26 paired, 60 unpaired) without ground-truths. Final submissions, including results, code, and factsheets were uploaded to Codabench \cite{coda}. Organizers independently verified the results, and top teams submitted training scripts to ensure reproducibility.

\paragraph{Ranking Statistics.} 
The final ranking (Table~\ref{tab:ntire_final_ranking}) is based on a weighted score (out of 100). To ensure fair weighting, raw metrics were mapped to $[0, 1]$ via Min-Max normalization across all teams. For metrics where higher is better (e.g., PSNR, SSIM, MoS, MUSIQUE, CLIPIA), the normalized score is $(x - x_{min}) / (x_{max} - x_{min})$. For metrics where lower is better (e.g., LPIPS, NIQUE), it is inverted as $(x_{max} - x) / (x_{max} - x_{min})$. Final scores depend on the track criteria. For the \textbf{Reference Track}, objective metrics are weighted equally; normalized PSNR, SSIM, and LPIPS each contribute $33.33\%$. For the \textbf{Non-Reference Track}, human visual preference is the gold standard, so the Mean Opinion Score (MoS) accounts for $50.00\%$, while NIQUE, MUSIQUE, and CLIPIA equally share the remaining half ($16.67\%$ each).

\begin{table*}[t]
\centering
\caption{Final rankings and evaluation metrics for the NTIRE 2026 LLIE (Track 1) and JDLLIE (Track 2) Challenges. The best results in each column per track are highlighted in \textbf{\textcolor{red}{red}}, and the second-best results are \underline{\textcolor{blue}{underlined blue}}. Note: Reference and Non-Reference tracks are ranked independently.}
\label{tab:ntire_final_ranking}

% Define custom commands for styling
\newcommand{\best}[1]{\textbf{\textcolor{red}{#1}}}
\newcommand{\second}[1]{\underline{\textcolor{blue}{#1}}}

% We use @{\hskip 4em} to create a wide visual gap between the Ref and Non-Ref sides
\resizebox{\textwidth}{!}{
\begin{tabular}{l ccccc @{\hskip 4em} l cccccc}
\toprule
\multicolumn{6}{c}{\textbf{Reference Track}} & \multicolumn{7}{c}{\textbf{Non-Reference Track}} \\
\cmidrule(lr){1-6} \cmidrule(lr){7-13}
\textbf{Team} & \textbf{PSNR} $\uparrow$ & \textbf{SSIM} $\uparrow$ & \textbf{LPIPS} $\downarrow$ & \textbf{Score} $\uparrow$ & \textbf{Rank} & \textbf{Team} & \textbf{MoS} $\uparrow$ & \textbf{NIQUE} $\downarrow$ & \textbf{MUSIQUE} $\uparrow$ & \textbf{CLIPIA} $\uparrow$ & \textbf{Score} $\uparrow$ & \textbf{Rank} \\
\midrule
\multicolumn{13}{c}{\textbf{Track 1: Low-Light Image Enhancement (LLIE)}} \\
\midrule
\rowcolor{gray!15} \textit{SYSU-FVL} & \second{20.71} & \best{0.6773} & 0.4802 & \best{94.80} & \textit{1} & \textit{AAIR-ARM} & 7.30 & \second{4.6992} & \second{31.62} & \second{0.3752} & \best{80.83} & \textit{1} \\
\midrule
WHU-MVP           & 19.75 & \second{0.6729} & 0.4880 & \second{87.77} & 2  & SNUCV             & 5.90 & 4.8965  & \best{32.18} & \best{0.4216} & \second{73.83} & 2  \\
KLETech-CEVI      & 19.96 & 0.6706 & 0.5340 & 84.59 & 3  & BAU-Vision        & \best{9.00} & 6.7578  & 24.80 & 0.1977 & 61.47  & 3  \\
BITssvgg          & \best{20.74} & 0.6482 & 0.5955 & 80.28 & 4  & PSU               & \second{7.40} & 6.4359  & 28.36 & 0.2484 & 59.64  & 4  \\
BAU-Vision        & 18.99 & 0.6308 & 0.4572 & 79.52 & 5  & TranssionAI       & 7.00 & 6.2586  & 26.58 & 0.2874 & 58.21  & 5  \\
SNUCV             & 18.00 & 0.6328 & \second{0.4242} & 76.93 & 6  & SYSU-FVL          & 6.80 & 6.5289  & 26.11 & 0.3060 & 56.23  & 6  \\
YuFans            & 17.81 & 0.6527 & 0.4993 & 72.28 & 7  & NTR               & 6.85 & 6.6029  & 27.89 & 0.2724 & 56.02  & 7  \\
AAIR-ARM          & 18.81 & 0.5649 & \best{0.4231} & 71.29 & 8  & YuFans            & 6.75 & 6.2365  & 27.61 & 0.2436 & 54.54  & 8  \\
TranssionAI       & 17.40 & 0.6532 & 0.4887 & 70.86 & 9  & WHU-MVP           & 6.65 & 5.8908  & 26.68 & 0.2465 & 54.49  & 9  \\
ReagvisLabs       & 18.48 & 0.6337 & 0.5269 & 70.85 & 10 & ReagvisLabs       & 6.20 & 5.7119  & 26.38 & 0.2541 & 52.25  & 10 \\
DH-XHDL-Team      & 18.60 & 0.6310 & 0.5641 & 67.83 & 11 & KLETech-CEVI      & 7.10 & 7.6378  & 25.90 & 0.2775 & 50.73  & 11 \\
NTR               & 16.82 & 0.6371 & 0.5323 & 61.14 & 12 & BITssvgg          & 7.20 & 7.4701  & 24.18 & 0.2700 & 49.51  & 12 \\
VesperLux         & 18.07 & 0.5350 & 0.5060 & 55.04 & 13 & MC2               & 4.80 & \best{4.1567}  & 23.47 & 0.3023 & 49.35  & 13 \\
PSU               & 15.07 & 0.6012 & 0.5303 & 45.45 & 14 & DH-XHDL-Team      & 6.25 & 6.7755  & 25.53 & 0.2696 & 47.59  & 14 \\
SOMIS-LAB         & 15.32 & 0.5642 & 0.5368 & 40.66 & 15 & VesperLux         & 5.50 & 5.7577  & 20.09 & 0.2328 & 37.32  & 15 \\
MC2               & 16.53 & 0.4616 & 0.5757 & 28.45 & 16 & SOMIS-LAB         & 2.50 & 7.3967  & 26.98 & 0.2763 & 19.60  & 16 \\
AnanyaHariniLaksh & 17.05 & 0.4957 & 0.8003 & 16.94 & 17 & AnanyaHariniLaksh & 2.15 & 11.1732 & 19.11 & 0.2729 & -11.33 & 17 \\
\midrule
\multicolumn{13}{c}{\textbf{Track 2: Joint Denoising and Low-Light Image Enhancement (JDLLIE)}} \\
\midrule
\rowcolor{gray!15} \textit{BAU-Vision} & \best{20.84} & 0.6789 & \best{0.4426} & \best{98.70} & \textit{1} & \textit{DH-XHDL-Team} & \best{8.50} & 7.1471 & 24.88 & 0.2686 & \best{75.34} & \textit{1} \\
\midrule
MiVideoDLLIE      & \second{20.75} & 0.6693 & 0.4739 & \second{94.02} & 2  & MC2               & 4.00 & \best{3.5335} & \best{26.86} & \best{0.4634} & \second{68.97} & 2  \\
DH-XHDL-Team      & 19.87 & \best{0.6964} & 0.4777 & 90.46 & 3  & BAU-Vision        & \second{8.25} & 7.2961 & 23.64 & 0.1920 & 65.79 & 3  \\
APRIL-AIGC        & 18.94 & \second{0.6831} & \second{0.4434} & 87.81 & 4  & WIRNet            & 5.00 & 6.6177 & 24.74 & \second{0.3085} & 54.56 & 4  \\
RetinexDualV2     & 18.69 & 0.6474 & 0.4999 & 77.46 & 5  & APRIL-AIGC        & 6.00 & 7.7581 & 22.67 & 0.3068 & 54.09 & 5  \\
Lucky one         & 20.23 & 0.6658 & 0.6011 & 76.62 & 6  & MiVideoDLLIE      & 5.50 & \second{5.8237} & 23.10 & 0.2610 & 53.22 & 6  \\
NTR               & 17.95 & 0.6537 & 0.5474 & 68.40 & 7  & RetinexDualV2     & 5.75 & 7.4865 & 23.31 & 0.2420 & 50.45 & 7  \\
VesperLux         & 17.94 & 0.5949 & 0.5402 & 64.78 & 8  & VesperLux         & 6.05 & 8.7387 & 21.88 & 0.2628 & 47.72 & 8  \\
WIRNet            & 17.35 & 0.5682 & 0.5002 & 63.76 & 9  & weichow           & 4.00 & 6.6101 & \second{26.21} & 0.2261 & 45.91 & 9  \\
PSU TEAM          & 17.18 & 0.5946 & 0.5385 & 60.47 & 10 & PSU TEAM          & 4.50 & 7.1230 & 23.53 & 0.2705 & 44.91 & 10 \\
KLETech-CEVI      & 17.17 & 0.5960 & 0.5739 & 56.60 & 11 & SOMIS-LAB         & 3.25 & 6.9525 & 23.51 & 0.2601 & 35.98 & 11 \\
SOMIS-LAB         & 16.75 & 0.5694 & 0.5913 & 50.30 & 12 & Lucky one         & 5.25 & 10.8288 & 19.43 & 0.3040 & 34.46 & 12 \\
weichow           & 15.65 & 0.5639 & 0.6070 & 41.71 & 13 & KLETech-CEVI      & 1.25 & 8.0173 & 23.00 & 0.2580 & 18.49 & 13 \\
MC2               & 15.13 & 0.2457 & 0.7440 & 0.00  & 14 & NTR               & 2.00 & 8.7089 & 21.93 & 0.2206 & 17.38 & 14 \\
\bottomrule
\end{tabular}
}
\end{table*}

\section{Challenge Methods}
\label{subsec:methods}

\begin{disclaimerbox}
In the following Sections, we describe the top challenge solutions -- each was checked manually by the organizers to ensure fairness.

Note that the method descriptions were provided by each team as their contribution to this report.
\end{disclaimerbox}

% \newpage

\subsection{RLLIE: Robust Low-light Image Enhancement}
\label{subsec:team01-SYSU-FVL}

%%%%%%%%%%%%%%%%%%%%%%%%%%%%%%%%%%%%
% \begin{center}

% \vspace{2mm}
% \noindent\emph{\textbf{RLLIE: Robust Low-light Image Enhancement}}
% \vspace{2mm}

% \noindent\emph{Zhi Jin\textsuperscript{1,2,3}, Hongjun Wu\textsuperscript{1}, Wenjian Zhang\textsuperscript{1}, Chang Ye\textsuperscript{1}}

% \vspace{2mm}

% \noindent\emph{1.School of Intelligent Systems Engineering, Shenzhen Campus of Sun Yat-sen University, Shenzhen, 518107, China}

% \noindent\emph{2.Guangdong Provincial Key Laboratory of Fire Science and Intelligent Emergency Technology, Shenzhen, 518107, China}

% \noindent\emph{3.Guangdong Provincial Key Laboratory of Robotics and Digital Intelligent Manufacturing Technology, Guangzhou, 510535, China
% }
% \end{center}

%%%%%%%%%%%%%%%%%%%%%%%%%%%%%%%%%%%%%%%%%%%%%%%%%%%%%%%%%%%%%%%%%%

\paragraph{Method.}
% As illustrated in Fig.\ref{fig:RUHDLLIE}, 
\emph{SYSU-FVL}'s proposed enhancement framework employs two distinct models: ESDNet \cite{esdnet} and DarkIR \cite{darkir}. Notably, both models utilize a U-Net-like hierarchical network architecture, which endows the enhancement results with superior robustness. These two networks are trained independently. During inference, they perform a weighted fusion of their individual outputs to achieve optimal performance. Specifically, the fusion weights for ESDNet ($\alpha$) and DarkIR ($\beta$) are empirically set to 0.66 and 0.34, respectively.

The total training loss $\mathcal{L}_{total}$ is computed as follows: 
$$\mathcal{L}_{total} =   \mathcal{L}_C(I_{pred}, I_{gt}) + \lambda \mathcal{L}_P(I_{pred}, I_{gt})$$,
where $I_{pred}$ and $I_{gt}$ denote the predicted image and the corresponding ground truth, respectively. $\mathcal{L}_C$ represents the Charbonnier loss \cite{chan}, and $\mathcal{L}_P$ denotes the perceptual loss extracted using a pre-trained VGG-19 network\cite{vgg19}. The balancing hyperparameter $\lambda$ is empirically set to 0.04.

% \begin{figure}[htbp]
% \begin{center}
% \includegraphics[width=0.47\textwidth]{teams/imgs/SYSU-FVL-RLIE.pdf}
% \end{center}
% \vspace{-1em}
% \caption{Pipeline of RLLIE proposed by \emph{SYSU-FVL}}
% \label{fig:RUHDLLIE}
% \vspace{-1em}
% \end{figure}

% \paragraph{Training Details.}
% Inspired by MIRNetv2 \cite{mirnetv2}, they adopt a progressive training strategy for their models. Both ESDNet and DarkIR are trained from scratch using the Adam optimizer \cite{adam}. The initial learning rate is set to $2\times10^{-4}$ and is regulated by a cyclic cosine annealing schedule \cite{cosine}. During the initial training phase, they set the batch size to 8 and randomly crop $512 \times 512$ image patches. In the subsequent fine-tuning stage, paired images from the validation set are incorporated into the training, and the learning rate is reduced to $2\times10^{-6}$.

% For ESDNet, the model is trained for a total of 291,500 iterations. The initial phase lasts for 290,000 iterations. The fine-tuning stage is divided into two steps: 1,000 iterations with a batch size of 4 and a patch size of 1024, followed by a final 500 iterations with a batch size of 1 and a patch size of 2048.

% For DarkIR, the model undergoes 301,500 iterations in total during training. After 300,000 iterations in the initial phase, the model enters the fine-tuning stage for the remaining 1,500 iterations, where the batch size is set to 2, and the patch size is increased to 1024.
\subsection{LL-ESDNet: Learning Hierarchical Coarse-to-Fine Image Enhancement}
\label{subsec:team02-WHU-MVP}

%%%%%%%%%%%%%%%%%%%%%%%%%%%%%%%%%%%%
% \begin{center}

% \vspace{2mm}
% \noindent\emph{\textbf{LL-ESDNet: Learning Hierarchical Coarse-to-Fine Image Enhancement}}
% \vspace{2mm}

% \noindent\emph{Xunpeng Yi, Qinglong Yan, Yibing Zhang}

% \vspace{2mm}

% \noindent\emph{Wuhan University}

% \end{center}

%%%%%%%%%%%%%%%%%%%%%%%%%%%%%%%%%%%%%%%%%%%%%%%%%%%%%%%%%%%%%%%%%%

\paragraph{Method.}
\emph{WHU-MVP} introduced LL-ESDNet, which follows a paradigm in which an initial enhancement is first performed and subsequently leveraged as guidance for refined restoration. First, a Transformer-based network is employed to obtain an initial enhancement in the downsampled resolution space. This result is subsequently used as guidance for further refinement and enhancement. The refinement backbone is inspired by ESDNet~\cite{yu2022towards} and incorporates cross-resolution mutual attention with the coarse enhancement results during the encoding stage. Meanwhile, an adaptive resolution alignment mechanism is adopted within this process. By imposing joint constraints on both the coarse enhancement and refinement networks, and integrating hierarchical attention guidance with a progressive enhancement strategy, the final output is achieved.

% \paragraph{Training Details.} They utilize the full-resolution images from the LSD dataset provided by the organizers, rather than the cropped patch version. Training is performed exclusively on four NVIDIA RTX 4090 GPUs. A progressive training strategy is adopted, where the patch sizes are set to ${960, 1280, 1280, 1440, 1600}$, with corresponding batch sizes of ${4, 2, 2, 2, 1}$, and training iterations of ${36\text{K}, 24\text{K}, 12\text{K}, 28\text{K}, 50\text{K}}$, respectively. The initial learning rate is set to $3 \times 10^{-4}$ scheduled using cyclic cosine annealing.

\subsection{UHDM: Universal Hierarchical Decomposition Model for Low-Light Image Enhancement}
\label{subsec:team03-KLETech-CEVI}

%%%%%%%%%%%%%%%%%%%%%%%%%%%%%%%%%%%%
% \begin{center}

% \vspace{2mm}
% \noindent\emph{\textbf{Universal Hierarchical Decomposition Model for LLIE}}
% \vspace{2mm}

% \noindent\emph{Zaynab Ali$^1$, Saiprasad Meesiyawar$^3$, Nikhil Akalwadi$^3$, Ramesh Ashok Tabib$^{2,3}$, Uma Mudenagudi$^{2,3}$}

% \vspace{2mm}

% \noindent\emph{$^1$Department of Computer Applications, $^2$ Department of Electronics and Communication Engineering, $^3$ Center for Visual Intelligence (CEVI), KLE Technological University, Hubballi, INDIA}

% \end{center}

%%%%%%%%%%%%%%%%%%%%%%%%%%%%%%%%%%%%%%%%%%%%%%%%%%%%%%%%%%%%%%%%%%

\paragraph{Method.}
\emph{KLETech-CEVI} utilizes the \textbf{UHDM} architecture,
% ilustrated in Figure \ref{fig:team03-kletech-cevi}, 
which employs a hierarchical encoder--decoder framework to capture multi-scale feature representations. This design progressively extracts features ($F_i = E_i(x)$) that encode both global illumination and fine spatial structures, enabling effective enhancement while maintaining structural consistency. The decoder reconstructs the final image by aggregating these hierarchical features:
\begin{equation}
\hat{y} = D(F_1, F_2, \ldots, F_n)
\end{equation}
\textbf{Loss Function:}
To balance reconstruction accuracy and perceptual quality, the team optimizes the model using a weighted multi-term loss function:
\begin{equation}
L = L_{char} + 0.1L_{ssim} + 0.15L_{edge}
\end{equation}
where $L_{char}$ represents the Charbonnier loss, $L_{ssim}$ ensures structural similarity, and $L_{edge}$ preserves fine edge details.

% \begin{figure*}[!t]
% \centering
% \includegraphics[width=\linewidth]{teams/team03-kletech-cevi.jpeg}
% \vspace{-6pt}
% \caption{Overview of the proposed UHDM architecture by \emph{KLETech-CEVI}}
% \vspace{-10pt}
% \label{fig:team03-kletech-cevi}
% \end{figure*}

% \paragraph{Training Details.} The proposed model is trained using Python and PyTorch Framework with settings as mention in Table \ref{table:kletech-cevi}.

% \begin{table}[!hbtp]
% \centering
% \caption{Technical summary of UHDM model proposed by Team KLETech-CEVI.}
% \vspace{-2pt}
% \small
% \setlength{\tabcolsep}{6pt}
% \renewcommand{\arraystretch}{1.2}
% \resizebox{\columnwidth}{!}{% % Added resizebox here
% \begin{tabular}{lcccccccc}
% \hline
% Input & Time & Iter. & Extra & Diff. & Attn. & Quant. & Params (M) & Runtime \\
% \hline
% 384×384 & 27h & 290k & No & No & Yes & No & 41.5 & 1.2 s \\
% \hline
% \end{tabular}
% } % End of resizebox
% \vspace{-6pt}\label{table:kletech-cevi}
% \end{table}
% \subsection{DERNet: Histoformer crossed with a dynamic expert mechanism}
\subsection{DERNet}
\label{subsec:team04-BITssvgg}

%%%%%%%%%%%%%%%%%%%%%%%%%%%%%%%%%%%%
% \begin{center}

% \vspace{2mm}
% \noindent\emph{\textbf{Contribution Title}}
% \vspace{2mm}

% \noindent\emph{Author 1, Author 2}

% \vspace{2mm}

% \noindent\emph{Affiliation Lab}

% \end{center}

%%%%%%%%%%%%%%%%%%%%%%%%%%%%%%%%%%%%%%%%%%%%%%%%%%%%%%%%%%%%%%%%%%

% \textcolor{red}{sho:}

\paragraph{Method.}

\emph{BITssvgg}'s proposed solution, termed DERNet
% , of which architecture is depicted in Figure \ref{fig:bitssvgg}, 
is a dedicated encoder-decoder framework for low-light image enhancement that jointly models local degradation patterns and long-range structural dependencies. Built upon Histoformer \cite{one1} and a dynamic expert mechanism \cite{two2}, the architecture combines Transformer blocks for global contextual modeling with dynamic expert routing units to handle spatially varying degradation. In the encoder, Transformer blocks capture long-range interactions, while a routing branch dynamically aggregates multiple lightweight convolutional experts for content-adaptive local restoration. To keep computational overhead manageable, each expert adopts a compact design consisting of a $1 \times 1$ pointwise convolution, a $3 \times 3$ depthwise convolution, a GELU nonlinearity, and another $1 \times 1$ pointwise convolution. By predicting expert-wise weights conditioned on the current feature map, the network adapts to regions with varying illumination and texture characteristics. Finally, the decoder employs a Transformer-based architecture for progressive global reconstruction, effectively improving brightness and local contrast while preserving fine textures and structural consistency.
\subsection{Wave-P: Wavelet Feature Propagation in Learnable Polar Color Space}
\label{subsec:team05-BAU-Vision}

%%%%%%%%%%%%%%%%%%%%%%%%%%%%%%%%%%%%
% \begin{center}

% \vspace{2mm}
% \noindent\emph{\textbf{Wavelet Feature Propagation in HVI Color Space}}
% \vspace{2mm}

% \noindent\emph{Furkan Kınlı}

% \vspace{2mm}

% \noindent\emph{Bahçeşehir University}

% \end{center}

%%%%%%%%%%%%%%%%%%%%%%%%%%%%%%%%%%%%%%%%%%%%%%%%%%%%%%%%%%%%%%%%%%

\paragraph{Method.}
\emph{BAU-Vision} introduces \textit{Wave-P},
% described in Figure \ref{fig:bau-arch}, 
an architecture that improves low-light restoration by combining sub-band focused feature propagation with global light modulation. While the \textit{CIDNet} baseline using the HVI color space \cite{yan2025hvi} effectively decouples luminance and chrominance, its reliance on standard interpolation-based sampling leads to significant high-frequency information loss. Their primary contribution, \textit{Wavelet-based feature propagation}, replaces lossy downsampling and upsampling layers with Discrete Wavelet Transforms (DWT) and Inverse Wavelet Transforms (IWT) \cite{liu2018multi}. By decomposing each feature map into $LL, LH, HL,$ and $HH$ sub-bands and concatenating them, the encoder preserves textural details across the multi-scale hierarchy. This anti-aliased propagation is critical for recovering structural realism for the extreme exposures and residual sensor noise typical of low-light scenarios.

To ensure global consistency across the enhanced scene, the team introduces a Global Light Modulation (GLM) at the architectural bottleneck. The GLM utilizes global average pooling followed by a sequential convolutional block with SiLU activation \cite{elfwing2018sigmoid} to generate a global scaling factor. This mechanism dynamically modulates local feature responses based on holistic scene illumination, preventing over-exposure or localized artifacts in heterogeneous lighting conditions. Interaction between the Intensity ($I$) and Chromatic ($H,V$) streams follows the CIDNet dual-path structure \cite{yan2025hvi}, but mediated by multiple Lightweight Cross-Attention (LCA) blocks, where the intensity path guides the chrominance restoration via temperature-scaled attention.

% \begin{figure*}[htbp]
% \begin{center}
% \includegraphics[width=\linewidth]{figures/BAU-arch.png}
% \end{center}
% \vspace{-1em}
% \caption{\emph{BAU-Vision}'s Wave-P architecture}
% \label{fig:bau-arch}
% \vspace{-1em}
% \end{figure*}

% \paragraph{Training Details.} To stabilize training on given challenge dataset, They define a sample-adaptive coefficient vector $\alpha \in \mathbb{R}^N$ for each batch of $N$ samples. For the $i$-th sample, the coefficient $\alpha_i$ is computed as the maximum ratio between the ground-truth and predicted global mean intensities ($\mu$): $\alpha_i = \max(\mu_{\text{GT}}^i / \mu_{\text{pred}}^i, \mu_{\text{pred}}^i / \mu_{\text{GT}}^i)$. This formulation balances gradient contributions across the batch regardless of absolute brightness levels, thus preventing optimization bias toward high-intensity samples. The network is supervised by a dual-domain multi-objective loss that combines Charbonnier \cite{charbonnier1997deterministic}, SSIM, and Laplacian edge terms with \textit{VGG-19} based perceptual similarity \cite{johnson2016perceptual} to enforce joint pixel-level and structural consistency. The optimization is performed using the AdamW \cite{loshchilov2017decoupled} regulated by cosine annealing scheduler. To enhance model robustness against diverse lighting conditions, they incorporate random gamma adjustments ($\gamma \in [0.6, 1.2]$) during training and employ 6-view TTA during the inference.
% \subsection{MB-LPFR: CIDNet and OSEDiff fusion refined by ORNet}
\subsection{MB-LPFR}
\label{subsec:team06-SNUCV}

%%%%%%%%%%%%%%%%%%%%%%%%%%%%%%%%%%%%
% \begin{center}

% \vspace{2mm}
% \noindent\emph{\textbf{MB-LPFR : Multi-Branch Low-Light Image Enhancement via Laplacian Pyramid Fusion and Refinement}}
% \vspace{2mm}

% \noindent\emph{Donghun Ryou, Inju Ha, Junoh Kang, Bohyung Han}

% \vspace{2mm}

% \noindent\emph{Seoul National University}

% \end{center}

%%%%%%%%%%%%%%%%%%%%%%%%%%%%%%%%%%%%%%%%%%%%%%%%%%%%%%%%%%%%%%%%%%
% \begin{figure}[t]
% \centering
% \includegraphics[width=0.45\textwidth]{teams/snucv_overview.pdf}
% \caption{\emph{SNUCV}'s MB-LPFR pipeline}
% \label{fig:MB-LPFR}
% \end{figure}

\paragraph{Method.}
% Figure~\ref{fig:MB-LPFR} illustrates the 
\emph{SNUCV}'s overall pipeline consists of three stages. First, CIDNet+~\cite{yan2025hvi} and OSEDiff~\cite{wu2024one} independently process the input image. While CIDNet+ processes the entire image at once, OSEDiff processes it patch by patch due to its large model size. To prevent visible grid artifacts caused by per-patch brightness inconsistencies, they explicitly train OSEDiff to accept a globally downsampled version of the input image as an additional condition alongside each local patch. This global conditioning is then consistently applied during inference to ensure uniform brightness across all patches. Furthermore, to enhance OSEDiff's performance, they train it using an advanced image-conditioned manifold regularization method (ICM-SR~\cite{kang2025icm}). Second, multi-scale fusion is performed via a three-level Laplacian pyramid: low-frequency components (overall tone) are extracted from the CIDNet+ output, while high-frequency components (fine details and textures) are derived from the OSEDiff output. Finally, a refinement step is applied using ORNet~\cite{ryou2025beyond}.

% \paragraph{Training Details.}
% They trained CIDNet+~\cite{yan2025hvi} and OSEDiff~\cite{wu2024one} exclusively on the LSD training dataset~\cite{sharif2026illuminating}. For CIDNet+, they followed the original paper's protocol and cropped patches to a size of $1280 \times 1280$. OSEDiff was initialized from a pre-trained Stable Diffusion checkpoint (CompVis/stable-diffusion-v1-4) and fine-tuned on the LSD dataset according to the ICM-SR~\cite{kang2025icm} training procedure. For ORNet~\cite{ryou2025beyond}, they directly utilized the publicly released pre-trained weights without any additional fine-tuning.

\subsection{BAPE: Brightness-Aware Progressive Enhancement}
\label{subsec:team07-YuFans}

%%%%%%%%%%%%%%%%%%%%%%%%%%%%%%%%%%%%
% \begin{center}

% \vspace{2mm}
% \noindent\emph{\textbf{Brightness-Aware Progressive Enhancement (BAPE)}}
% \vspace{2mm}

% \noindent\emph{Wei Zhou, Linfeng Li, Hongyuan Huang}

% \vspace{2mm}

% \noindent\emph{National University of Singapore, Zhejiang University}

% \end{center}

%%%%%%%%%%%%%%%%%%%%%%%%%%%%%%%%%%%%%%%%%%%%%%%%%%%%%%%%%%%%%%%%%%

\paragraph{Method.}
\emph{YuFans} presents BAPE, a data-centric training methodology that addresses the critical \textit{brightness distribution gap} between the LSD training data and competition test images. Competition images (mean intensity $\approx$17.5/255) are 2--3$\times$ darker than available training sources (mean $\approx$28--58/255), making domain adaptation---not architectural design---the primary challenge. Evidence: CIDNet~\cite{yan2025hvi} (CVPR 2025) achieved only 13.85\,dB in zero-shot, \textit{worse} than naive gamma correction (14.91\,dB).

Their approach
% , illustrated in Fig.~\ref{fig:bape_pipeline}, 
is built on NAFNet~\cite{chen2022nafnet} (width=64, 116M parameters) and comprises three synergistic techniques:

\textbf{(1) Brightness-Aware Augmentation (BAA):} They apply stochastic gamma correction ($\gamma \sim \mathcal{U}[0.7, 2.5]$) and brightness jittering \textit{exclusively to input images} during training, simulating extreme low-light conditions without modifying ground truth targets.

\textbf{(2) Progressive Patch Escalation (PPE):} A multi-stage fine-tuning schedule ($384 \rightarrow 512 \rightarrow 768$) progressively increases spatial context, reducing the patch-to-full-image PSNR gap from 3.7\,dB to 0.75\,dB.

\textbf{(3) Multi-Source Domain Bridging (MSDB):} The team curates 200 full-resolution pairs from four LSD In-the-wild subsets (DLI, DLO, DEI, DEO), spanning brightness levels 21--42. The DEI/DEO subsets, closest to competition brightness, serve as domain bridges, reducing overfitting vs.\ using only 24 validation pairs.

During inference, they use tiled processing ($768\!\times\!768$, overlap 128, Hanning window blending) with $8\times$ geometric TTA (4~rotations $\times$ 2~flips). LSD pretraining contributes +4.51\,dB, PPE adds +0.60\,dB, and MSDB improves generalization (independent val 24.54\,dB vs.\ competition val 23.79\,dB, a gap of only 0.75\,dB).

\subsection{LFM-LLIE: Latent Flow-Matching Framework for Low-Light Image Enhancement}
\label{subsec:team08-AAIR_ARM}

%%%%%%%%%%%%%%%%%%%%%%%%%%%%%%%%%%%%
% \begin{center}

% \vspace{2mm}
% \noindent\emph{\textbf{Latent Flow-Matching for Low-Light Image Enhancement}}
% \vspace{2mm}

% \noindent\emph{Yuval Haitman, Ariel Lapid, Reuven Peretz, Idit Diamant}

% \vspace{2mm}

% \noindent\emph{Applied AI Research (AAIR), Arm Ltd}

% \end{center}

% \begin{figure}[ht]
% \centering
% \includegraphics[
%     width=0.9\linewidth,
%     trim=0cm 0cm 4cm 0cm,
%     clip
% ]{teams/team08-AAIR_ARM_figure.pdf}
% \caption{\textbf{\emph{AAIR\_ARM}'s LFM-LLIE overview.} \textbf{(a) Training:} A flow-matching network $u_\theta$ (HDiT-based) is trained in latent space using interpolated samples $z_t$ created from the noise-perturbed low-light latent image $\tilde{z}_1$. \textbf{(b) Inference:} Starting from $z_1 = \mathcal{E}(x_1)$, the latent is iteratively transformed by the learned velocity field $u_\theta$ into $\hat{z}_0$ and decoded by $\mathcal{D}$ to produce the enhanced image $\hat{x}_0$.}
% \label{fig:aairarm}
% \end{figure}

%%%%%%%%%%%%%%%%%%%%%%%%%%%%%%%%%%%%%%%%%%%%%%%%%%%%%%%%%%%%%%%%%%

\paragraph{Method.}
\emph{AAIR-ARM} proposes \textit{\textbf{LFM-LLIE}}
% (Figure \ref{fig:aairarm})
, a latent flow-matching framework for low-light image enhancement that learns a mapping from the distribution of low-light images to that of normal-light images in a compact latent space. 
A key design choice is operating in \textit{latent space}, which reduces computational cost while enabling modeling of semantically meaningful image statistics~\cite{stablediff}. In addition, they adopt \textit{HDiT} as the backbone of $u_\theta$, leveraging its transformer-based architecture with large receptive fields and skip connections~\cite{hdit,ohayon2025posteriormean} to effectively capture global illumination and fine textures, resulting in realistic and high-quality enhancements.
Given a low-light image $x_1$, a frozen VAE encoder $\mathcal{E}$ maps it to a latent representation $z_1$. A flow-matching network $u_\theta$ is then trained to transport low-light latents toward their clean counterparts $z_0$. Following~\cite{lipman2023flowmatchinggenerativemodeling}, they learn the conditional velocity field using interpolated samples $z_t = (1 - t)z_0 + t\tilde{z}_1$, where $\tilde{z}_1 = z_1 + \sigma\epsilon$, $\epsilon \sim \mathcal{N}(0, I)$, and $t \sim \mathcal{U}[0,1]$. The noise-perturbed latent $\tilde{z}_1$ improves robustness and stabilizes training~\cite{cohen2025efficient}. At inference, the learned velocity field $u_\theta$ is used to iteratively solve the ODE with $M$ steps, mapping $z_1$ to a restored latent $\hat{z}_0$, which is decoded by the VAE decoder $\mathcal{D}$ to produce the enhanced image $\hat{x}_0$.

% \paragraph{Training Details.}
% They use a frozen Stable Diffusion VAE \cite{stablediff} and an HDiT-XL/4 backbone~\cite{hdit} to parameterize the flow model. The network is trained from scratch on paired low-normal light images from the Smartphone Dataset (LSD)~\cite{sharif2026illuminating}. Training is performed on $2048 \times 2048$ image patches (randomly cropped from the 4K images) with a batch size of 8 for 80K optimization steps using AdamW. The model is optimized with the standard flow-matching objective on the conditional velocity field~\cite{lipman2023flowmatchinggenerativemodeling}, with noise level $\sigma=0.01$. All experiments are conducted using PyTorch on 8 NVIDIA H100 GPUs. At inference, the ODE is solved using 25 function evaluations (NFE).
\subsection{Progressive patch training strategy for TFFormer}
\label{subsec:team09-TranssionAI}

%%%%%%%%%%%%%%%%%%%%%%%%%%%%%%%%%%%%
% \begin{center}

% \vspace{2mm}
% \noindent\emph{\textbf{Contribution Title}}
% \vspace{2mm}

% \noindent\emph{Author 1, Author 2}

% \vspace{2mm}

% \noindent\emph{Affiliation Lab}

% \end{center}

% \begin{figure}[ht]
% \centering
% \includegraphics[width=0.9\linewidth]{teams/imgs/pipeline.png}
% \caption{\emph{TranssionAI} overview}
% \label{fig:transsitionai}
% \end{figure}

%%%%%%%%%%%%%%%%%%%%%%%%%%%%%%%%%%%%%%%%%%%%%%%%%%%%%%%%%%%%%%%%%%

% \textcolor{red}{COPY-PASTED:}

\paragraph{Method.}

% To effectively train TFFormer \cite{sharif2026illuminating} on high-resolution images while maintaining training stability and GPU memory efficiency, \emph{TranssionAI} adopts a progressive patch training strategy. Specifically, instead of directly training the model on large patches, they gradually increase the training patch size in multiple stages, as shown in Figure \ref{fig:transsitionai}. The training procedure starts from smaller patches to allow the model to quickly learn local illumination and texture priors. As training progresses, larger patches are introduced to enable the network to capture long-range spatial dependencies and global illumination relationships. This progressive strategy provides two key benefits: Smaller patches allow faster convergence and more stable optimization in early stages. Larger patches gradually introduce broader spatial context, which is critical for low-light enhancement tasks. In practice, the model weights from the previous stage are used to initialize the next stage.
To effectively train TFFormer \cite{sharif2026illuminating} on high-resolution images while maintaining training stability and GPU memory efficiency, \emph{TranssionAI} adopts a progressive patch training strategy. Specifically, instead of directly training the model on large patches, they gradually increase the training patch size in multiple stages
% , as shown in Figure \ref{fig:transsitionai}
. The training procedure starts from smaller patches to allow the model to quickly learn local illumination and texture priors. As training progresses, larger patches are introduced to enable the network to capture long-range spatial dependencies and global illumination relationships. This progressive strategy provides two key benefits: Smaller patches allow faster convergence and more stable optimization in early stages. Larger patches gradually introduce broader spatial context, which is critical for low-light enhancement tasks. In practice, the model weights from the previous stage are used to initialize the next stage.

\subsection{Weighted Late-Fusion Ensemble of LLIEMoE and RetinexNet}
\label{subsec:team10-ReagvisLabs}

%%%%%%%%%%%%%%%%%%%%%%%%%%%%%%%%%%%%
% \begin{center}

% \vspace{2mm}
% \noindent\emph{\textbf{Weighted Late-Fusion Ensemble of LLIEMoE and RetinexNet for In-the-Wild Low-Light Image Enhancement}}
% \vspace{2mm}

% \noindent\emph{Prateek Shaily$^{1}$, Jayant Kumar$^{1}$, Hardik Sharma$^{2}$, Ashish Negi$^{2}$, Sachin Chaudhary$^{3}$, Akshay Dudhane$^{4}$ ,Praful Hambarde$^{2}$, Amit Shukla$^{2}$}

% \vspace{2mm}

% \noindent\emph{$^{1}$Reagvis Labs Pvt. Ltd.}
% \noindent\emph{$^{2}$Indian Institute of Technology Mandi}
% \noindent\emph{$^{3}$COE:AI, School of Computer Science, UPES Dehradun, India}
% \noindent\emph{$^{4}$Mohamed bin Zayed University of Artificial Intelligence, Abu Dhabi}

% \end{center}

%%%%%%%%%%%%%%%%%%%%%%%%%%%%%%%%%%%%%%%%%%%%%%%%%%%%%%%%%%%%%%%%%%

\paragraph{Method.}
\emph{ReagvisLabs} presents a two-branch low-light enhancement framework for the NTIRE~2026 low-light image enhancement track~\cite{ntire2026lowlight}. The final prediction is obtained through weighted late fusion of two complementary models: a spatial Mixture-of-Experts branch, denoted as LLIEMoE, and an illumination-aware Retinex-based branch derived from RetinexFormer~\cite{cai2023retinexformer}. Given a low-light input image $x$, the two branches produce enhanced outputs $\hat{y}_{\mathrm{moe}}$ and $\hat{y}_{\mathrm{ret}}$, which are combined as
\begin{multline}
\hat{y} = w_{\mathrm{moe}} \hat{y}_{\mathrm{moe}} + w_{\mathrm{ret}} \hat{y}_{\mathrm{ret}},
\quad
(w_{\mathrm{moe}}, w_{\mathrm{ret}}) = \\ = (0.500458, 0.499542).
\end{multline}
The LLIEMoE branch is built on a shared encoder and three specialized decoders, following the principle of expert routing in Mixture-of-Experts models~\cite{shazeer2017moe}. These experts are designed to focus on fidelity preservation, perceptual detail recovery, and naturalness restoration, respectively. The naturalness expert incorporates Retinex-style reflectance--illumination recombination~\cite{wei2018retinex} together with channel recalibration inspired by squeeze-and-excitation attention~\cite{hu2018se}. A lightweight gating module predicts spatially varying expert weights from darkness, texture, and chromatic cues, enabling content-adaptive restoration across different image regions. In parallel, the RetinexNet branch follows a multi-stage illumination estimation and illumination-guided restoration design derived from RetinexFormer~\cite{cai2023retinexformer}, which strengthens brightness recovery and denoising in severely underexposed areas. The final ensemble therefore combines the local specialization capability of expert routing with the global illumination modeling strength of Retinex-based restoration.

\subsection{TEI-LLIE}
\label{subsec:team11-DH-XHDL-Team}

%%%%%%%%%%%%%%%%%%%%%%%%%%%%%%%%%%%%
% \begin{center}

% \vspace{2mm}
% \noindent\emph{\textbf{Towards Efficient and Illumination-Robust Ultra-High-Definition Low-Light Image Enhancement}}
% \vspace{2mm}

% \noindent\emph{MoHao Wu, Lin Wang}

% \vspace{2mm}

% \noindent\emph{Zhejiang Dahua Technology Co.,Ltd.}

% \end{center}

%%%%%%%%%%%%%%%%%%%%%%%%%%%%%%%%%%%%%%%%%%%%%%%%%%%%%%%%%%%%%%%%%%

\paragraph{Method.}
\emph{DH-XHDL-Team}'s solution adopts a dedicated encoder-decoder architecture for low-light image enhancement\cite{yu2022efficientscalerobustultrahighdefinitionimage}
% , as Figure \ref{fig:placeholder} displays
. The framework first employs a pre-processing head that downsamples the input image by a factor of 2 via pixel shuffle, followed by a 5×5 convolutional layer to extract low-level features and expand the receptive field. The core backbone consists of an encoder-decoder structure with three downsampling/upsampling stages, integrated with skip connections to enable high-resolution feature reuse and facilitate high-definition image restoration. At each decoder stage, intermediate outputs are generated through convolution and pixel shuffle upsampling, supervised by ground-truth images for deep training.\\The network incorporates a novel customized modules: \textbf{Linear-Focused Network Block (LFNblock):} Replaces the CA and GELU layers in PlainNet with SCA and SimpleGate, reducing computational redundancy and multiplication operations, mitigating gradient vanishing caused by low-intensity pixels (critical for low-light scenarios), and enhancing performance with fewer parameters.
\subsection{Pretraining and Finetuning of MDAE}
\label{subsec:team12-NTR}

%%%%%%%%%%%%%%%%%%%%%%%%%%%%%%%%%%%%
% \begin{center}

% \vspace{2mm}
% \noindent\emph{\textbf{Supervised Fine-Tuning of a Masked-Diffusion Autoencoder for Low-Light Image Enhancement}}
% \vspace{2mm}

% \noindent\emph{Jiachen Tu, Guoyi Xu, Yaoxin Jiang, Jiajia Liu, Yaokun Shi}

% \vspace{2mm}

% \noindent\emph{University of Illinois Urbana-Champaign}

% \end{center}

%%%%%%%%%%%%%%%%%%%%%%%%%%%%%%%%%%%%%%%%%%%%%%%%%%%%%%%%%%%%%%%%%%

\paragraph{Method.}
\emph{NTR}'s approach consists of two stages: (1)~self-supervised pretraining with Masked-Diffusion Autoencoders (MDAE)~\cite{tu2026mdae}, and (2)~supervised fine-tuning on the low-light enhancement task.

\textbf{Pretraining.}
MDAE learns image representations by simultaneously applying two complementary corruptions to clean images: spatial patch masking and diffusion noise.
Given a clean image $X_0$, MDAE first adds Gaussian noise $\tilde{X}_t = X_0 + \sigma_t Z$ and then zeros out randomly selected patches to produce a doubly-corrupted input $\tilde{X}_t^M = M_v \odot \tilde{X}_t$.
The network $g_\theta(\tilde{X}_t^M, t)$ is trained to predict the clean image $X_0$ from this input, supervised by two losses: a masked region reconstruction loss (learning global structure by inpainting hidden content) and a visible region denoising loss (learning fine-grained texture by removing noise).
The noise level $t$ is injected at every network stage via FiLM conditioning~\cite{perez2018film}, enabling the model to adapt its reconstruction strategy across the corruption spectrum.
This dual-corruption objective interpolates between masked autoencoding (as $\sigma_t \to 0$) and denoising score matching (as masking ratio $\to 0$), learning representations that capture both structural coherence and textural detail within a single pretraining objective.

\textbf{Architecture.}
The backbone is TimeDiffiT-ResNet-128~\cite{tu2025score}, a U-Net-style encoder-decoder with 142.5\,M parameters.
It follows a four-stage hierarchy with channel multipliers $(1, 2, 4, 8)$ relative to a base width of 128.
Each stage contains two ResNet blocks with group normalization and a time-conditioned self-attention layer.
The diffusion timestep modulates each block through FiLM conditioning: $h_{\text{out}} = h_{\text{in}} \odot (\gamma(t_{\text{emb}}) + 1) + \beta(t_{\text{emb}})$, where $\gamma$ and $\beta$ are learned projections from a sinusoidal time embedding.
Skip connections link each encoder stage to its corresponding decoder stage.

\textbf{Supervised fine-tuning.}
After MDAE pretraining on large-scale natural images, they fine-tune the model on the low-light enhancement task using paired data.
During fine-tuning, the time-conditioning input is fixed at $\sigma{=}0$, effectively converting the noise-conditioned model into a deterministic feed-forward restorer while retaining the multi-scale representations learned during pretraining.

\subsection{ISALux-ModalFormer Network}
\label{subsec:team13-VesperLux}

%%%%%%%%%%%%%%%%%%%%%%%%%%%%%%%%%%%%
% \begin{center}

% \vspace{2mm}
% \noindent\emph{\textbf{VesperLux: Hi/Lo-Fi Modulation Vision Transformer Using Hybrid Deep Supervision For Low Light Image Enhancement}}
% \vspace{2mm}

% \noindent\emph{Raul Balmez, Alexandru Brateanu, Ciprian Orhei, Codruta Ancuti, Cosmin Ancuti}

% \vspace{2mm}

% \noindent\emph{University of Manchester / Polytehnic University of Timisoara}

% \end{center}

%%%%%%%%%%%%%%%%%%%%%%%%%%%%%%%%%%%%%%%%%%%%%%%%%%%%%%%%%%%%%%%%%%

\paragraph{Method.}
\emph{VesperLux}'s network follows a U-shaped architecture with Transformer-based processing blocks \cite{balmez2026isalux, brateanu2025modalformer}. The encoder progressively downsamples the input through two stages using strided convolutions, while the decoder reconstructs the full-resolution output via transposed convolutions with skip connections. Each processing block consists of a channel-wise self-attention module followed by a feed-forward network (FFN), with residual connections and layer normalization applied at each stage.

The attention mechanism operates in the channel dimension \cite{zamir2022restormer}. Given input $\mathbf{x} \in \mathbb{R}^{B \times C \times H \times W}$, queries, keys and values are obtained via a depthwise separable projection \cite{balmez2025depthlux}, then reshaped to $\mathbf{Q}, \mathbf{K}, \mathbf{V} \in \mathbb{R}^{B \times h \times d \times HW}$, where $h$ is the number of heads and $d = C/h$. Attention is computed as:
\begin{equation}
    \text{Attn}(\mathbf{Q}, \mathbf{K}, \mathbf{V}) = \text{softmax}\!\left(\hat{\mathbf{Q}}\hat{\mathbf{K}}^\top \cdot \tau\right)\mathbf{V},
\end{equation}
where $\hat{\mathbf{Q}}, \hat{\mathbf{K}}$ are $\ell_2$-normalized and $\tau$ is a learned temperature per head.

Encoder blocks additionally incorporate a Hi-Fi/Lo-Fi modulation module. The feature map is decomposed into low-frequency ($\mathbf{x}_\text{lo}$) and high-frequency ($\mathbf{x}_\text{hi} = \mathbf{x} - \mathbf{x}_\text{lo}$) components via average pooling, each independently modulated and recombined:
\begin{equation}
    \mathbf{x}_\text{mod} = \text{Conv}\!\left[\sigma(\mathbf{x}_\text{hi}) \,\|\, \sigma(\mathbf{x}_\text{lo})\right],
\end{equation}
where $\sigma$ denotes GELU activation and $[\,\|\,]$ denotes channel concatenation.

The network produces three outputs: the main full-resolution prediction $\hat{y}$, a mid-decoder auxiliary output $\hat{y}_\text{mid}$, and a bottleneck auxiliary output $\hat{y}_\text{bn}$ (both upsampled to input resolution). The total loss is:
\begin{equation}
    \mathcal{L} = \lambda_1 \mathcal{L}_\text{MSE}(\hat{y}, y) + \lambda_2 \mathcal{L}_\text{color}(\hat{y}_\text{mid}, y) + \lambda_3 \mathcal{L}_\text{light}(\hat{y}_\text{bn}, y),
\end{equation}
with weights $\lambda_1{=}1.0$, $\lambda_2{=}0.5$, $\lambda_3{=}0.25$. The color loss penalizes angular deviation between predicted and target pixel colors:
\begin{equation}
    \mathcal{L}_\text{color} = \mathcal{L}_1 + \mathbb{E}\!\left[1 - \frac{\hat{y} \cdot y}{\|\hat{y}\|\|y\|}\right].
\end{equation}
The light loss operates in log space on the max-RGB channel $v = \max_c(\cdot)$ to focus on exposure:
\begin{equation}
    \mathcal{L}_\text{light} = \mathcal{L}_1\!\left(\log(\hat{v} + \epsilon),\, \log(v + \epsilon)\right).
\end{equation}

% \paragraph{Training Details.}
% The model is trained for $300\text{k}$ iterations using the Adam optimizer ($\beta_1{=}0.9$, $\beta_2{=}0.999$) with an initial learning rate of $2{\times}10^{-4}$ and gradient clipping. The learning rate schedule consists of two phases: a constant phase for the first $92\text{k}$ iterations at $3{\times}10^{-4}$, followed by cosine annealing over the remaining $208\text{k}$ iterations from $3{\times}10^{-4}$ to $10^{-6}$. Data augmentation includes MixUp with $\beta{=}1.2$, retaining the identity sample with a fixed probability.
\subsection{DUSKAN: Dual Spectral Kolmogorov-Arnold Network}
\label{subsec:team14-PSU}

%%%%%%%%%%%%%%%%%%%%%%%%%%%%%%%%%%%%
% \begin{center}

% \vspace{2mm}
% \noindent\emph{\textbf{DUSKAN: Dual Spectral Kolmogorov-Arnold Network}}
% \vspace{2mm}

% \noindent\emph{Bilel Benjdira, Anas M. Ali, Wadii Boulila}

% \vspace{2mm}

% \noindent\emph{Robotics and Internet of Things Laboratory (RIOTU Lab), Prince Sultan University, Riyadh 11586, Saudi Arabia}

% \end{center}
%%%%%%%%%%%%%%%%%%%%%%%%%%%%%%%%%%%%%%%%%%%%%%%%%%%%%%%%%%%%%%%%%%

\paragraph{Method.}
Effective image restoration requires capturing both \emph{global} degradation patterns (severe underexposure, residual noise spectra, colour shifts) and fine \emph{local} texture. \emph{PSU} addresses this with \textbf{DUSKAN}, a dual-path network whose two branches are purpose-built for complementary representations: spectral-spatial processing captures global frequency structure alongside local spatial detail, while Kolmogorov-Arnold adaptive processing learns flexible, data-driven nonlinearities that classical fixed activations cannot express. Both paths operate in parallel and are blended with a learned gating weight, allowing the network to emphasise whichever representation is most useful at each scale. The backbone is a symmetric 4-level U-Net with global residual learning; every stage is a \textsc{DUSKANBlock}
 % (Fig.~\ref{fig:duskan_arch})
.

\noindent\textbf{Path A: Spectral-Spatial Processing.}
Low-level degradations manifest as structured changes in the Fourier spectrum. Given features $\mathbf{X}{\in}\mathbb{R}^{B\times C\times H\times W}$, the \emph{spectral branch} computes the 2D FFT $\hat{\mathbf{X}}{=}\mathcal{F}(\mathbf{X})$, separates magnitude $|\hat{\mathbf{X}}|$ from phase $\angle\hat{\mathbf{X}}$, and enriches the magnitude with: \textit{(i)} separable learnable positional encodings, \textit{(ii)} SE channel reweighting~\cite{senet2018}, and \textit{(iii)} a LayerNorm-MLP refinement. The refined magnitude is recombined with the original phase, inverse-transformed, and scaled by a learnable sigmoid gate, yielding global features $\mathbf{X}_{\mathrm{spec}}$. Complementing this, the \emph{spatial branch} uses $1{\times}1$ expansion to $2C$ channels, parallel $3{\times}3$ and $5{\times}5$ depthwise convolutions, sigmoid gating, and $1{\times}1$ projection back to $C$. Both branches are fused via a two-layer $1{\times}1$ convolution bottleneck:
\begin{equation}
  \mathbf{Y}_A = f_{\mathrm{fuse}}\!\bigl( \sigma(w_s)\,\mathbf{X}_{\mathrm{spat}} \;\|\; \sigma(w_f)\,\mathbf{X}_{\mathrm{spec}}\bigr)
\end{equation}
where $f_{\mathrm{fuse}} = \mathrm{Conv}\!\to\!\mathrm{GELU}\!\to\!\mathrm{Conv}$.

\noindent\textbf{Path B: Kolmogorov-Arnold Adaptive Processing.}
Inspired by the Kolmogorov-Arnold representation theorem~\cite{kan2025}, they replace every linear projection with a \emph{KolmogorovArnoldLinear} layer that learns its own activation function per edge via polynomial basis expansion. Inside the selective gated mixer, after KAN projection, 1D depthwise convolution, and SiLU activation, output features are computed as a parallel-additive gate:
\begin{equation}
  \mathbf{y} = \mathbf{x}_m \odot \sigma\!\bigl(\mathrm{softplus}(\mathbf{W}_g\mathbf{x}_m)\bigr) + \mathbf{x}_m \odot \mathbf{d}
\end{equation}
where $\mathbf{d}$ is a learnable skip scale ensuring a stable residual gradient path. 

\noindent\textbf{DUSKANBlock and Network.}
Outputs $\mathbf{Y}_A$ and $\mathbf{Y}_B$ are blended by a single real-valued logit $\alpha$ trained independently per stage: $\mathbf{Y} = \sigma(\alpha)\,\mathbf{Y}_A + \bigl(1{-}\sigma(\alpha)\bigr)\,\mathbf{Y}_B$. Both paths follow a pre-norm residual pattern with near-zero initialization~\cite{nafnet2022}. The U-Net uses strided $2{\times}2$ convolution downsampling, PixelShuffle upsampling~\cite{pixelshuffle}, and a global input-to-output residual. Training minimises $\mathcal{L} = \mathcal{L}_1 + 0.01\,\mathcal{L}_{\mathrm{VGG}} + 0.1\,\mathcal{L}_{\mathrm{FF}}$, combining pixel-level fidelity, semantic fidelity via VGG-19 features~\cite{vggloss}, and spectral fidelity via focal frequency loss~\cite{ffl2021}.

% \begin{figure*}[t]
%   \centering
%   \includegraphics[width=\textwidth]{teams/duskan_ntire2_resized.png}
%   \caption{\textbf{DUSKAN architecture by \emph{PSU}.} \textit{Top}: Symmetric 4-level U-Net with DUSKANBlock stages, strided downsampling, PixelShuffle upsampling, and global residual learning. \textit{Bottom}: DUSKANBlock detail. \textbf{Path A} (blue) extracts global features via FFT magnitude modulation and fuses them with local multi-scale depthwise convolution features. \textbf{Path B} (red) uses Kolmogorov-Arnold polynomial-basis activations with a parallel-additive selective gate.}
%   \label{fig:duskan_arch}
% \end{figure*}

% \paragraph{Training Details.} 
% The network (45.6M parameters) was trained from scratch for 500 epochs with a batch size of 2 on a single NVIDIA A100 (80GB) GPU. Training utilized $512 \times 512$ random patches with geometric augmentations (random rotations and flips). They employed the AdamW optimizer ($\beta_1 = 0.9, \beta_2 = 0.999$, weight decay $= 10^{-4}$) with an initial learning rate of $2 \times 10^{-4}$, gradually decreased to $1 \times 10^{-6}$ via a Cosine Annealing scheduler. PyTorch's Automatic Mixed Precision (AMP) with FP16 gradients was used to maximize efficiency, resulting in a training time of roughly 49 hours. During inference, an overlapping patch-based strategy was applied to reconstruct the full-resolution test images.
\subsection{APDNet: Asymmetric Physics-Guided Dual-Domain Network for Low-Light Image Enhancement}
\label{subsec:team15-SOMIS-LAB}

%%%%%%%%%%%%%%%%%%%%%%%%%%%%%%%%%%%%
% \begin{center}

% \vspace{2mm}
% \noindent\emph{\textbf{Asymmetric Physics-Guided Dual-Domain Network for Low-Light Image Enhancement}}
% \vspace{2mm}

% \noindent\emph{Kaifan Qiao, Bofei Chen}

% \vspace{2mm}

% \noindent\emph{Beihang University}

% \end{center}

%%%%%%%%%%%%%%%%%%%%%%%%%%%%%%%%%%%%%%%%%%%%%%%%%%%%%%%%%%%%%%%%%%

\paragraph{Method.}

\emph{SOMIS-LAB} proposes APDNet, an Asymmetric Physics-Guided Dual-Domain Network tailored for low-light image enhancement. Inspired by LEDNet \cite{zhou2022lednet} and DarkIR \cite{feijoo2025darkir}, APDNet adopts a encoder-decoder architecture. However, the encoder and decoder are uniquely designed to accommodate the joint processing of multiple tasks. Furthermore, an illumination decoupling module is introduced prior to the encoder to extract the global illumination information from the input image. Specifically, the encoder takes both the low-light image and the global illumination information as inputs to perform illumination and color restoration. To further enhance the restoration efficacy, they incorporate the wavelet transform to process the features in the frequency domain. Conversely, the decoder focuses on detail restoration, operating in the spatial domain and utilizing the low-light image as input. Previous asymmetric models often incorporated intermediate losses to deliberately segregate the functionalities of the encoder and the decoder. However, the team argues that such a rigid division is suboptimal, and they empirically observe that this practice leads to performance degradation. Consequently, APDNet dispenses with intermediate losses, relying instead on a unified holistic loss to drive the model toward superior performance.

% \paragraph{Training Details.}

% To ensure stable training on the provided challenge dataset, They crop the original images into $512\times512$ patches. The network is optimized using the AdamW optimizer with an initial learning rate of $ 3e-4$  for 200 epochs. To enhance the generalization capability of the model, data augmentation techniques, including random horizontal flipping and random rotations, are applied during the training phase. All experiments are implemented using the PyTorch framework and conducted on an NVIDIA RTX 3090 GPU. Our comprehensive loss function comprises an L1 loss, a perceptual loss, and an SSIM loss, ensuring that the enhanced images maintain high fidelity with the ground truth references at pixel, perceptual, and structural levels.

% \subsection{DNDiff: training-free Diffusion-Retinex++ denoised by MCSCNet}
\subsection{DNDiff}
\label{subsec:team16-MC2}

%%%%%%%%%%%%%%%%%%%%%%%%%%%%%%%%%%%%
% \begin{center}

% \vspace{2mm}
% \noindent\emph{\textbf{Denoising Enhanced Diffusion-Retinex for Low-Light Image Enhancement}}
% \vspace{2mm}

% \noindent\emph{Jingyi Xu, Xin Deng, Mai Xu, Shengxi Li, Lai Jiang}

% \vspace{2mm}

% \noindent\emph{MC2 Lab,  the School of Electronic and Information Engineering, Beihang University}

% \end{center}

%%%%%%%%%%%%%%%%%%%%%%%%%%%%%%%%%%%%%%%%%%%%%%%%%%%%%%%%%%%%%%%%%%

\paragraph{Method.}
% \emph{MC2} adopts Diffusion-Retinex++ \cite{yi2025diff} as our baseline for low-light enhancement, followed by MCSCNet \cite{xu2022revisiting} for post-denoising. To address the extreme low-light conditions and severe noise characteristic of Track 2, they perform adaptive denoising on the enhanced outputs. This approach effectively mitigates illumination-induced artifacts and noise, yielding a PSNR gain of over 1 dB. They intentionally avoid denoising as a preprocessing step, as raw low-light noise is complex and difficult to model using simple Gaussian or Poisson distributions, particularly given the lack of paired training data. Conversely, the residual noise in the enhanced images partially aligns with the Gaussian distribution, allowing the pre-trained MCSCNet to effectively eliminate it. As depicted in Figure \ref{fig:DNDiff}, DNDiff yields results with superior color fidelity, striking a natural balance between vibrant tones and structural accuracy.
\emph{MC2} adopts Diffusion-Retinex++ \cite{yi2025diff} as our baseline for low-light enhancement, followed by MCSCNet \cite{xu2022revisiting} for post-denoising. To address the extreme low-light conditions and severe noise characteristic of Track 2, they perform adaptive denoising on the enhanced outputs. This approach effectively mitigates illumination-induced artifacts and noise, yielding a PSNR gain of over 1 dB. They intentionally avoid denoising as a preprocessing step, as raw low-light noise is complex and difficult to model using simple Gaussian or Poisson distributions, particularly given the lack of paired training data. Conversely, the residual noise in the enhanced images partially aligns with the Gaussian distribution, allowing the pre-trained MCSCNet to effectively eliminate it. 
% As depicted in Figure \ref{fig:DNDiff},
DNDiff yields results with superior color fidelity, striking a natural balance between vibrant tones and structural accuracy.

% \begin{figure}[htbp]
% \begin{center}
% \includegraphics[width=0.47\textwidth]{teams/imgs/MC2.pdf}
% \end{center}
% \vspace{-1em}
% \caption{Qualitative comparison of DNDiff's color performance for low-light enhancement proposed by \emph{MC2}.}
% \label{fig:DNDiff}
% \vspace{-1em}
% \end{figure}

% \paragraph{Training Details.}
% Notably, the DNDiff is a training-free approach, which bypasses the need for extensive retraining on specific noise distributions.

\subsection{Fine-tuned MBLLEN}
\label{subsec:team17-AnanyaHariniLaksh}

%%%%%%%%%%%%%%%%%%%%%%%%%%%%%%%%%%%%
% \begin{center}

% \vspace{2mm}
% \noindent\emph{\textbf{Contribution Title}}
% \vspace{2mm}

% \noindent\emph{Author 1, Author 2}

% \vspace{2mm}

% \noindent\emph{Affiliation Lab}

% \end{center}

%%%%%%%%%%%%%%%%%%%%%%%%%%%%%%%%%%%%%%%%%%%%%%%%%%%%%%%%%%%%%%%%%%

% \textcolor{red}{COPY-PASTED:}

\paragraph{Method.}

\emph{AnanyaHariniLaksh} used MBLLEN \cite{mbllen} (Multi Branch Low Light Enhancement Network) a pre-trained CNN based Low Light Image Enhancement architecture. The team fine-tuned the pre-trained model.

% \subsection{DNATT: TFFormer-based optimization}
\subsection{DNATT}
\label{subsec:team18-Lucky-one}

%%%%%%%%%%%%%%%%%%%%%%%%%%%%%%%%%%%%
% \begin{center}

% \vspace{2mm}
% \noindent\emph{\textbf{Denoising Enhanced Tuning-fork-shaped Attention based network for Low-Light Image Enhancement}}
% \vspace{2mm}

% \noindent\emph{Jingyi Xu, Ying Xu, Xinyi Zhu}

% \vspace{2mm}

% \noindent\emph{Beihang University, Beijing Zhonghai Industry Company, Beijing Dongsi Shisi Tiao Primary School}

% \end{center}

%%%%%%%%%%%%%%%%%%%%%%%%%%%%%%%%%%%%%%%%%%%%%%%%%%%%%%%%%%%%%%%%%%

\paragraph{Method.}

\emph{Lucky-one} takes TFFormer \cite{sharif2026illuminating} as the baseline model for the low-light image enhancement task. Aiming at the characteristics of extremely low-light and high-noise images in Track 2, they optimize the original modules of TFFormer, such as the IG\_MSA\_M module, to enable effective denoising and filtering with a negligible increase in model parameters.
% The architecture of DNATT is shown in Fig. \ref{fig:DNATT}.

% \begin{figure*}[htbp]
% \begin{center}
% \includegraphics[width=\textwidth]{teams/imgs/Luckyone.pdf}
% \end{center}
% \vspace{-1em}
% \caption{DNATT pipeline of Team Lucky one.}
% \label{fig:DNATT}
% \vspace{-1em}
% \end{figure*}

% \paragraph{Training Details.}
% They introduce customized data augmentation strategies to retrain the improved network, employing a learning rate of 1e-4 to ensure stable and effective convergence. Trained on an NVIDIA GeForce RTX 4090 GPU. Furthermore, they re-calibrate the loss function weights to prioritize the restoration of Peak Signal-to-Noise Ratio (PSNR). The resulting composite loss function is defined as:
% \begin{equation}
%   \mathcal{L} = \mathcal{L}_{\text{MAE}} + \mathcal{L}_{\text{Luminance}},
% \end{equation}
% Additionally, an ensemble strategy employing a cross-model weighted average is incorporated during the inference phase, significantly enhancing the model's overall robustness and predictive accuracy.

\subsection{Exploring Large Diffusion Models for Joint Denoising and Low-Light Enhancement}
\label{subsec:team19-APRIL-AIGC}

%%%%%%%%%%%%%%%%%%%%%%%%%%%%%%%%%%%%
% \begin{center}

% \vspace{2mm}
% \noindent\emph{\textbf{Exploring Large Diffusion Models for Joint Denoising and Low-Light Enhancement}}
% \vspace{2mm}

% \noindent\emph{Shijun Shi, Jiangning Zhang, Yong Liu, Kai Hu, Jing Xu, Xianfang Zeng}

% \vspace{2mm}

% \noindent\emph{Jiangnan University, Zhejiang University, University of Science and Technology of China}

% \end{center}

%%%%%%%%%%%%%%%%%%%%%%%%%%%%%%%%%%%%%%%%%%%%%%%%%%%%%%%%%%%%%%%%%%

\paragraph{Method.}
\emph{APRIL-AIGC} adopts a single-stage restoration pipeline built on the FLUX.2-klein-4B-base diffusion model~\cite{flux-2-2025}. The backbone is optimized with image-conditioned fine-tuning under the rectified-flow formulation~\cite{liu2022flow}. Given a low-light input image, they first encode it into a condition latent $z_c$ with the VAE encoder. During denoising, the current latent state $z_t$ and the condition latent $z_c$ are patchified and concatenated before entering the multimodal diffusion transformer: $H = [\mathcal{P}(z_t); \mathcal{P}(z_c)]$, where $\mathcal{P}(\cdot)$ denotes latent patchification and token packing.

The main practical issue is the resolution gap between $512 \times 512$ training crops and full-resolution test images. Dense sliding-window inference introduces visible illumination seams near tile boundaries, especially with larger overlaps or more windows. They therefore adopt a fixed $2 \times 3$ tiling layout with moderate overlap and merge the restored tiles after diffusion sampling. 
\subsection{MiDLLIE}
\label{subsec:team20-MiVideoDLLIE}

%%%%%%%%%%%%%%%%%%%%%%%%%%%%%%%%%%%%
% \begin{center}

% \vspace{2mm}
% \noindent\emph{\textbf{MiDLLIE: A joint Denoising and Lowlight Enhancement Method}}
% \vspace{2mm}

% \noindent\emph{Jinao Song, GuangSheng Tang, Cheng Li, Yuqiang Yang, \\ Ziyi Wang, Yan Chen, Long Bao, Heng Sun}

% \vspace{2mm}

% \noindent\emph{Xiaomi Inc., China}

% \end{center}

%%%%%%%%%%%%%%%%%%%%%%%%%%%%%%%%%%%%%%%%%%%%%%%%%%%%%%%%%%%%%%%%%%

\paragraph{Method.}

\emph{MiVideoDLLIE}'s method adopts a two-stage model architecture: a "denoising and low-light enhancement" architecture. In the first stage, NAFNet \cite{chen2022simple} is used to distill the pre-trained Restormer \cite{zamir2022restormer} denoising model for effective noise removal. In the second stage, MoCEIR \cite{zamfir2025complexity} is employed to perform low-light enhancement.
% The pipeline of MiDLLIE is shown in Figure \ref{fig:MiDLLIE}.

% \begin{figure}[ht]
% \begin{center}
% \includegraphics[width=0.47\textwidth]{teams/imgs/MiVideoDLLIE.pdf}
% \end{center}
% \vspace{-1em}
% \caption{\emph{MiVideoDLLIE}'s pipeline of MiDLLIE}
% \label{fig:MiDLLIE}
% \vspace{-1em}
% \end{figure}

% \paragraph{Training Details.}

% During the training phase, to improve training efficiency while preserving image high-frequency details as much as possible, they resize the training data from $512\times512$ to $256\times256$. The optimizer is Adam with hyperparameters $\beta_1 = 0.9$ and $\beta_2 = 0.99$, and the initial learning rate is set to $1 \times 10^{-4}$, using a cosine annealing with restarts strategy. For the denoising stage, the NAFNet output is supervised by both the Restormer prediction and the ground truth via reconstruction loss and structural loss, with loss weights of 1 and 0.08, respectively. For the low-light enhancement stage, the MoCEIR output is optimized using reconstruction loss, frequency-domain loss, perceptual loss, and IQA loss against the ground truth, with respective weights of 1, 0.01, 0.02, and 0.02.

\subsection{RetinexDualV2: RetinexDual with explicit physical grounding}
% \subsection{RetinexDualV2}
\label{subsec:team21-RetinexDualV2}

%%%%%%%%%%%%%%%%%%%%%%%%%%%%%%%%%%%%
% \begin{center}

% \vspace{2mm}
% \noindent\emph{\textbf{RetinexDualV2: Physically-Grounded Dual Retinex for Generalized UHD Image Restoration}}
% \vspace{2mm}

% \noindent\emph{Mohab Kishawy, Jun Chen}

% \vspace{2mm}

% \noindent\emph{Department of Electrical and Computer Engineering, McMaster University}

% \end{center}

%%%%%%%%%%%%%%%%%%%%%%%%%%%%%%%%%%%%%%%%%%%%%%%%%%%%%%%%%%%%%%%%%%
% \begin{figure}
% \centering
% \includegraphics[width=\linewidth]{teams/RetinexDualV2.pdf} 
% \caption{RetinexDualV2 Overview.}
% \label{fig:RD_overview}
% \end{figure}

\paragraph{Method.}
% Team \emph{RetinexDualV2} proposes \textbf{RetinexDualV2} \cite{kishawy2026retinexdualv2physicallygroundeddualretinex}
% ,which advances its predecessor, RetinexDual \cite{kishawy2025retinexdualretinexbaseddualnature}, by introducing explicit physical grounding into the restoration pipeline. While retaining the foundational dual-branch framework for reflectance and illumination, the core novelty lies in the addition of a Task-Specific Physical Grounding Module (\textbf{TS-PGM}) as shown in Fig. \ref{fig:RD_overview}. Specifically for this challenge, the TS-PGM dynamically extracts degradation-aware, structure-aware illumination prior \cite{dong2025towards} directly from the noisy low-light input. Rather than conventional feature interactions, these physical vectors $P_m$ modulate feature processing via Physical-conditioned Multi-head Self-attention (\textbf{PC-MSA}). Given projected Query ($Q$), Key ($K$), and Value ($V$) representations, they modulate the value representation $V' = V \odot P_m$, computing the physical-conditioned attention output as:
Team \emph{RetinexDualV2} proposes \textbf{RetinexDualV2} \cite{kishawy2026retinexdualv2physicallygroundeddualretinex}
,which advances its predecessor, RetinexDual \cite{kishawy2025retinexdualretinexbaseddualnature}, by introducing explicit physical grounding into the restoration pipeline. While retaining the foundational dual-branch framework for reflectance and illumination, the core novelty lies in the addition of a Task-Specific Physical Grounding Module (\textbf{TS-PGM}). Specifically for this challenge, the TS-PGM dynamically extracts degradation-aware, structure-aware illumination prior \cite{dong2025towards} directly from the noisy low-light input. Rather than conventional feature interactions, these physical vectors $P_m$ modulate feature processing via Physical-conditioned Multi-head Self-attention (\textbf{PC-MSA}). Given projected Query ($Q$), Key ($K$), and Value ($V$) representations, they modulate the value representation $V' = V \odot P_m$, computing the physical-conditioned attention output as:
\begin{equation}
    \text{Attention}(Q, K, V') = \text{softmax}\left(\frac{K^{\top} Q}{\sigma}\right) V'
\end{equation}
where $\sigma$ rests as a learnable scale parameter. This physical integration explicitly guides the dual branches to guarantee physically plausible restoration.

\subsection{WIRNet: Gated LCA and DWT optimization for CIDNet}
% \subsection{WIRNet}
\label{subsec:team22-WIRNet}

%%%%%%%%%%%%%%%%%%%%%%%%%%%%%%%%%%%%
% \begin{center}

% \vspace{2mm}
% \noindent\emph{\textbf{Wavelet Intensity Refinement Network}}
% \vspace{2mm}

% \noindent\emph{Yi-Hao Cheng$^{1,3}$, Wan-Chi Siu (Leader)$^{1,2}$, Hon-Man Hammond Lee$^{1,2}$, and Chun-Chuen Hui$^{1,2}$}

% \vspace{2mm}

% \noindent\emph{$^{1}$Saint Francis University, $^{2}$The Hong Kong Polytechnic University, and $^{3}$The University of Hong Kong.}

% \end{center}

%%%%%%%%%%%%%%%%%%%%%%%%%%%%%%%%%%%%%%%%%%%%%%%%%%%%%%%%%%%%%%%%%%

\paragraph{Method.}

% Fig.~\ref{fig:WIRNet} shows our proposed Wavelet Intensity Refinement Network (WIRNet), which is built upon the CIDNet \cite{yan2025hvi} baseline. While CIDNet achieves competitive performance, it remains sensitive to noise and has limited ability to recover fine details. To address these limitations, we introduce three key contributions: 
% \begin{figure}[t]
% \centering
% \includegraphics[
%     width=0.6\linewidth
% ]{teams/team22-WIRNet_Figure.png}
% \caption{Overall Structure of WIRNet}
% \label{fig:WIRNet}
% \end{figure}

\textbf{\textit{Gated LCA module}}: Team \emph{WIRNet} designed a Gated Lighten Cross Attention (Gated LCA) module to adaptively weight the importance of features extracted from the attention branch, allowing the model to focus on information reconstruction and reduce noise disturbance.

% \textbf{\textit{Wavelet Residual Refinement block}}: The Wavelet Residual Refinement block is a frequency-aware module proposed for noisy low-light image enhancement. In CIDNet, the I branch on the left-hand side of Fig.~\ref{fig:WIRNet} can be considered as containing high-frequency and low-frequency information. Since noise is mostly high-frequency, the original CIDNet is unable to distinguish the noise, causing noise and high-frequency features to become entangled. Therefore, they introduced a Discrete Wavelet Transform (DWT) \cite{mallat2002theory, haar1911theorie} at the network’s bottleneck to decompose the I channel into low and high-frequency sub-bands, processed by two independent blocks respectively. This design allows the model to learn features without interference. Since the information at the bottleneck is concentrated, applying DWT at this stage ensures that the model can disentangle meaningful high-frequency details from noise more robustly and efficiently.

\textbf{\textit{Wavelet Residual Refinement block}}: The Wavelet Residual Refinement block is a frequency-aware module proposed for noisy low-light image enhancement. In CIDNet, the I branch can be considered as containing high-frequency and low-frequency information. Since noise is mostly high-frequency, the original CIDNet is unable to distinguish the noise, causing noise and high-frequency features to become entangled. Therefore, they introduced a Discrete Wavelet Transform (DWT) \cite{mallat2002theory, haar1911theorie} at the network’s bottleneck to decompose the I channel into low and high-frequency sub-bands, processed by two independent blocks respectively. This design allows the model to learn features without interference. Since the information at the bottleneck is concentrated, applying DWT at this stage ensures that the model can disentangle meaningful high-frequency details from noise more robustly and efficiently.

\textit{\textbf{Residual Aggregation Refinement}}: We added a Residual Aggregation Refinement block at the end of the network. This block effectively aggregates the deep features \cite{wang2020lightening} from both the I and HV branches to reconstruct the final high-quality output.

% \textcolor{red}{
Please refer to the \textbf{appendix \ref{sec:apd:details}} for implementation and training details of all participating methods.
% }
% \input{teams/team23-weichow}

\section{Discussion and Conclusion}
\label{sec:conclusion}
The NTIRE 2026 Challenge shifted low-light enhancement evaluation from synthesized to real-world data. Despite significant innovations, current methods still produce artifacts in extreme conditions, revealing a gap between benchmark metrics and real-world performance. Future editions will focus on closing this gap to reflect real-world complexities.

% In conclusion, the NTIRE 2026 Low Light Image Enhancement(LLIE) and joint denoising Challenges demonstrate significant progress in developing robust methods for enhancing low-light images. The diverse approaches and competitive results highlight the effectiveness of modern deep learning techniques in addressing noise and contrast degradation. Overall, the challenge provides valuable insights and benchmarks for future research in this field.
\section*{Acknowledgments}
% \textcolor{red}{
This work was partially supported by the Alexander von Humboldt Foundation. We thank the NTIRE 2026 sponsors: OPPO, Kuaishou, and University of W\"urzburg (Computer Vision Lab).
% }

{
    \small
    \bibliographystyle{ieeenat_fullname}

}

\clearpage

\appendix
\section*{Appendix}
\thispagestyle{empty}
\section{Teams and Affiliations}

\label{sec:apd:team}

\subsection*{Low Light Image Enhancement Challenge at NTIRE 2026}
\noindent\textit{\textbf{Members:}}\\ 
\textit{George Ciubotariu$^1$}, 
Sharif S M A$^2$
Abdur Rehman$^2$
Fayaz Ali Dharejo$^1$
Rizwan Ali Naqvi$^3$
Marcos V. Conde$^1$
Radu Timofte$^1$\\
\noindent\textit{\textbf{Affiliations: }}\\
$^1$ Computer Vision Lab, IFI \& CAIDAS, University of W\"urzburg \\
$^2$ Opt-AI Inc. \\
$^3$ Department of AI and Robotics, Sejong University, South Korea

% \subsection*{TemplateTeam}
% \noindent\textit{\textbf{Title: }}\\
% TemplateMethod\\
% \noindent\textit{\textbf{Members:}}\\ 
% \textit{TemplateLeader$^{1}$}, TemplateMember$^{1}$, TemplateMember$^{1}$, TemplateMember$^{1}$, TemplateMember$^{1}$\\
% \noindent\textit{\textbf{Affiliations: }}\\
% $^1$ TemplateInstitution1, $^2$ TemplateInstitution2\\

\subsection*{SYSU-FVL}
\noindent\textit{\textbf{Title: }}\\
RLLIE: Robust Low-light Image Enhancement\\
\noindent\textit{\textbf{Members:}}\\ 
\textit{Zhi Jin$^{1,2,3}$}, Hongjun Wu$^1$, Wenjian Zhang$^1$, Chang Ye$^1$\\
\noindent\textit{\textbf{Affiliations: }}\\
$^1$ .School of Intelligent Systems Engineering, Shenzhen Campus of Sun Yat-sen University, Shenzhen, 518107, China, $^2$ Guangdong Provincial Key Laboratory of Fire Science and Intelligent Emergency Technology, Shenzhen, 518107, China, $^3$ Guangdong Provincial Key Laboratory of Robotics and Digital Intelligent Manufacturing Technology, Guangzhou, 510535, China\\

\subsection*{WHU-MVP}
\noindent\textit{\textbf{Title: }}\\
LL-ESDNet: Learning Hierarchical Coarse-to-Fine Image Enhancement\\
\noindent\textit{\textbf{Members:}}\\ 
\textit{Xunpeng Yi$^{1}$}, Qinglong Yan$^{1}$, Yibing Zhang$^{1}$\\
\noindent\textit{\textbf{Affiliations: }}\\
$^1$ Wuhan University\\

% \subsection*{KLETech-CEVI}
% \noindent\textit{\textbf{Title: }}\\
% Universal Hierarchical Decomposition Model for LLIE\\
% \noindent\textit{\textbf{Members:}}\\ 
% \textit{Zaynab Ali$^1$}, Saiprasad Meesiyawar$^3$, Nikhil Akalwadi$^3$, Ramesh Ashok Tabib$^{2,3}$, Uma Mudenagudi$^{2,3}$\\
% \noindent\textit{\textbf{Affiliations: }}\\
% $^1$Department of Computer Applications, $^2$ Department of Electronics and Communication Engineering, $^3$ Center for Visual Intelligence (CEVI), KLE Technological University, Hubballi, INDIA\\
\subsection*{KLETech-CEVI}

\noindent\textit{\textbf{Title:}}\\
Universal Hierarchical Decomposition Model for LLIE

\noindent\textit{\textbf{Members:}}\\
\textit{
Zaynab Ali$^{1,2}$,
Saiprasad Meesiyawar$^{2}$,
Varda I Pattanshetty$^{2,3}$,
Varsha I Pattanshetty$^{2,3}$,
Nikhil Akalwadi$^{2}$,
Padmashree Desai$^{2,3}$,
Ramesh Ashok Tabib$^{2,4}$,
Uma Mudenagudi$^{2,4}$
}

\noindent\textit{\textbf{Affiliations:}}\\
$^1$ Department of Computer Applications, KLE Technological University, Hubballi, INDIA\\
$^2$ Center for Visual Intelligence (CEVI), KLE Technological University, Hubballi, INDIA\\
$^3$ School of Computer Science and Engineering, KLE Technological University, Hubballi, INDIA\\
$^4$ Department of Electronics and Communication Engineering, KLE Technological University, Hubballi, INDIA\\

\subsection*{BITssvgg}
\noindent\textit{\textbf{Title: }}\\
DERNet: Dynamic Expert Routing Network for Low-Light Image Enhancement\\
\noindent\textit{\textbf{Members:}}\\ 
\textit{Hao Yang$^{1}$}, Ruikun Zhang$^{1}$, Liyuan Pan$^{1}$\\
\noindent\textit{\textbf{Affiliations: }}\\
$^1$ Beijing Institute of Technology\\

\subsection*{BAU-Vision}
\noindent\textit{\textbf{Title: }}\\
Wavelet Feature Propagation in HVI Color Space\\
\noindent\textit{\textbf{Members:}}\\ 
\textit{Furkan Kınlı$^{1}$}\\
\noindent\textit{\textbf{Affiliations: }}\\
$^1$ Bahçeşehir University\\

\subsection*{SNUCV}
\noindent\textit{\textbf{Title: }}\\
MB-LPFR : Multi-Branch Low-Light Image Enhancement via Laplacian Pyramid Fusion and Refinement\\
\noindent\textit{\textbf{Members:}}\\ 
\textit{Donghun Ryou$^{1}$}, Inju Ha$^{1}$, Junoh Kang$^{1}$, Bohyung Han$^{1}$\\
\noindent\textit{\textbf{Affiliations: }}\\
$^1$ Seoul National University\\

\subsection*{YuFans}
\noindent\textit{\textbf{Title: }}\\
Brightness-Aware Progressive Enhancement (BAPE)\\
\noindent\textit{\textbf{Members:}}\\ 
\textit{Wei Zhou$^{1}$}, Linfeng Li$^{1}$, Hongyuan Huang$^{1}$\\
\noindent\textit{\textbf{Affiliations: }}\\
$^1$ National University of Singapore, Zhejiang University\\

\subsection*{AAIR-ARM}
\noindent\textit{\textbf{Title: }}\\
Latent Flow-Matching for Low-Light Image Enhancement\\
\noindent\textit{\textbf{Members:}}\\ 
\textit{Yuval Haitman$^{1}$}, Ariel Lapid$^{1}$, Reuven Peretz$^{1}$, Idit Diamant$^{1}$\\
\noindent\textit{\textbf{Affiliations: }}\\
$^1$ Applied AI Research (AAIR), Arm Ltd\\

\subsection*{TranssionAI}
\noindent\textit{\textbf{Title: }}\\
TFFormer with Progressive Multi-Scale Patch Training for Low-Light Image Enhancement\\
\noindent\textit{\textbf{Members:}}\\ 
\textit{Leilei Cao$^{1}$}, Shuo Zhang$^{1}$\\
\noindent\textit{\textbf{Affiliations: }}\\
$^1$ TEX AI, Transsion Holdings\\

\subsection*{ReagvisLabs}
\noindent\textit{\textbf{Title: }}\\
Weighted Late-Fusion Ensemble of LLIEMoE and RetinexNet for In-the-Wild Low-Light Image Enhancement\\
\noindent\textit{\textbf{Members:}}\\ 
\textit{Prateek Shaily$^{1}$}, Jayant Kumar$^{1}$, Hardik Sharma$^{2}$, Ashish Negi$^{2}$, Sachin Chaudhary$^{3}$, Akshay Dudhane$^{4}$ ,Praful Hambarde$^{2}$, Amit Shukla$^{2}$\\
\noindent\textit{\textbf{Affiliations: }}\\
$^{1}$Reagvis Labs Pvt. Ltd., $^{2}$Indian Institute of Technology Mandi, $^{3}$COE:AI, School of Computer Science, UPES Dehradun, India, $^{4}$Mohamed bin Zayed University of Artificial Intelligence, Abu Dhabi\\

\subsection*{DH-XHDL-Team}
\noindent\textit{\textbf{Title: }}\\
Towards Efficient and Illumination-Robust Ultra-High-Definition Low-Light Image Enhancement\\
\noindent\textit{\textbf{Members:}}\\ 
\textit{MoHao Wu$^{1}$}, Lin Wang$^{1}$\\
\noindent\textit{\textbf{Affiliations: }}\\
$^1$ Zhejiang Dahua Technology Co.,Ltd.\\

\subsection*{NTR}
\noindent\textit{\textbf{Title: }}\\
Supervised Fine-Tuning of a Masked-Diffusion Autoencoder for Low-Light Image Enhancement\\
\noindent\textit{\textbf{Members:}}\\ 
\textit{Jiachen Tu$^{1}$}, Guoyi Xu$^{1}$, Yaoxin Jiang$^{1}$, Jiajia Liu$^{1}$, Yaokun Shi$^{1}$\\
\noindent\textit{\textbf{Affiliations: }}\\
$^1$ University of Illinois Urbana-Champaign\\

\subsection*{VesperLux}
\noindent\textit{\textbf{Title: }}\\
Hi/Lo-Fi Modulation Vision Transformer Using Hybrid Deep Supervision For Low Light Image Enhancement\\
\noindent\textit{\textbf{Members:}}\\ 
\textit{Raul Balmez$^{1}$}, Alexandru Brateanu$^{1}$, Ciprian Orhei$^{2}$, Codruta Ancuti$^{2}$, Cosmin Ancuti$^{2}$\\
\noindent\textit{\textbf{Affiliations: }}\\
$^1$ University of Manchester, $^2$ Polytehnic University of Timisoara\\

\subsection*{PSU}
\noindent\textit{\textbf{Title: }}\\
DUSKAN: Dual Spectral Kolmogorov-Arnold Network\\
\noindent\textit{\textbf{Members:}}\\ 
\textit{Bilel Benjdira$^{1}$}, Anas M. Ali$^{1}$, Wadii Boulila$^{1}$\\
\noindent\textit{\textbf{Affiliations: }}\\
$^1$ Robotics and Internet of Things Laboratory (RIOTU Lab), Prince Sultan University, Riyadh 11586, Saudi Arabia\\

\subsection*{SOMIS-LAB}
\noindent\textit{\textbf{Title: }}\\
Asymmetric Physics-Guided Dual-Domain Network for Low-Light Image Enhancement\\
\noindent\textit{\textbf{Members:}}\\ 
\textit{Kaifan Qiao$^{1}$}, Bofei Chen$^{1}$\\
\noindent\textit{\textbf{Affiliations: }}\\
$^1$ Beihang University\\

\subsection*{MC2}
\noindent\textit{\textbf{Title: }}\\
Denoising Enhanced Diffusion-Retinex for Low-Light Image Enhancement\\
\noindent\textit{\textbf{Members:}}\\ 
\textit{Jingyi Xu$^{1}$}, Xin Deng$^{1}$, Mai Xu$^{1}$, Shengxi Li$^{1}$, Lai Jiang$^{1}$\\
\noindent\textit{\textbf{Affiliations: }}\\
$^1$ MC2 Lab, the School of Electronic and Information Engineering, Beihang University\\

\subsection*{AnanyaHariniLaksh}
\noindent\textit{\textbf{Title: }}\\
MBLLEN Fine-tuning\\
\noindent\textit{\textbf{Members:}}\\ 
\textit{Harini A$^{1}$}, Ananya N$^{1}$, Lakshanya K$^{1}$\\
\noindent\textit{\textbf{Affiliations: }}\\
$^1$ Shiv Nadar University Chennai, India\\

\subsection*{Lucky-one}
\noindent\textit{\textbf{Title: }}\\
Denoising Enhanced Tuning-fork-shaped Attention based network for Low-Light Image Enhancement\\
\noindent\textit{\textbf{Members:}}\\ 
\textit{Jingyi Xu$^{1}$}, Ying Xu$^{1}$, Xinyi Zhu$^{1}$\\
\noindent\textit{\textbf{Affiliations: }}\\
$^1$ Beihang University, Beijing Zhonghai Industry Company, Beijing Dongsi Shisi Tiao Primary School\\

\subsection*{APRIL-AIGC}
\noindent\textit{\textbf{Title: }}\\
Exploring Large Diffusion Models for Joint Denoising and Low-Light Enhancement\\
\noindent\textit{\textbf{Members:}}\\ 
\textit{Shijun Shi$^{1}$}, Jiangning Zhang$^{1}$, Yong Liu$^{1}$, Kai Hu$^{1}$, Jing Xu$^{1}$, Xianfang Zeng$^{1}$\\
\noindent\textit{\textbf{Affiliations: }}\\
$^1$ Jiangnan University, Zhejiang University, University of Science and Technology of China\\

\subsection*{MiVideoDLLIE}
\noindent\textit{\textbf{Title: }}\\
MiDLLIE: A joint Denoising and Lowlight Enhancement Method\\
\noindent\textit{\textbf{Members:}}\\ 
\textit{Jinao Song$^{1}$}, GuangSheng Tang$^{1}$, Cheng Li$^{1}$, Yuqiang Yang$^{1}$, Ziyi Wang$^{1}$, Yan Chen$^{1}$, Long Bao$^{1}$, Heng Sun$^{1}$\\
\noindent\textit{\textbf{Affiliations: }}\\
$^1$ Xiaomi Inc., China\\

% \subsection*{TemplateTeam}
% \noindent\textit{\textbf{Title: }}\\
% TemplateMethod\\
% \noindent\textit{\textbf{Members:}}\\ 
% \textit{TemplateLeader$^{1}$}, TemplateMember$^{1}$, TemplateMember$^{1}$, TemplateMember$^{1}$, TemplateMember$^{1}$\\
% \noindent\textit{\textbf{Affiliations: }}\\
% $^1$ TemplateInstitution1, $^2$ TemplateInstitution2\\

\subsection*{RetinexDualV2}
\noindent\textit{\textbf{Title: }}\\
Physically-Grounded Dual Retinex for Generalized UHD Image Restoration\\
\noindent\textit{\textbf{Members:}}\\ 
\textit{Mohab Kishawy$^{1}$}, Jun Chen$^{1}$\\
\noindent\textit{\textbf{Affiliations: }}\\
$^1$ Department of Electrical and Computer Engineering, McMaster University\\

\subsection*{WIRNet}
\noindent\textit{\textbf{Title: }}\\
Wavelet Intensity Refinement Network\\
\noindent\textit{\textbf{Members:}}\\ 
\textit{Yi-Hao Cheng$^{1,3}$}, Wan-Chi Siu (Leader)$^{1,2}$, Hon-Man Hammond Lee$^{1,2}$, and Chun-Chuen Hui$^{1,2}$\\
\noindent\textit{\textbf{Affiliations: }}\\
$^{1}$Saint Francis University, $^{2}$The Hong Kong Polytechnic University, and $^{3}$The University of Hong Kong.\\

\section{Implementation Details}
\label{sec:apd:details}

\subsection*{SYSU-FVL}

\paragraph{Training Details}
Inspired by MIRNetv2 \cite{mirnetv2}, they adopt a progressive training strategy for their models. Both ESDNet and DarkIR are trained from scratch using the Adam optimizer \cite{adam}. The initial learning rate is set to $2\times10^{-4}$ and is regulated by a cyclic cosine annealing schedule \cite{cosine}. During the initial training phase, they set the batch size to 8 and randomly crop $512 \times 512$ image patches. In the subsequent fine-tuning stage, paired images from the validation set are incorporated into the training, and the learning rate is reduced to $2\times10^{-6}$.

For ESDNet, the model is trained for a total of 291,500 iterations. The initial phase lasts for 290,000 iterations. The fine-tuning stage is divided into two steps: 1,000 iterations with a batch size of 4 and a patch size of 1024, followed by a final 500 iterations with a batch size of 1 and a patch size of 2048.

For DarkIR, the model undergoes 301,500 iterations in total during training. After 300,000 iterations in the initial phase, the model enters the fine-tuning stage for the remaining 1,500 iterations, where the batch size is set to 2, and the patch size is increased to 1024.

\begin{figure}[htbp]
\begin{center}
\includegraphics[width=0.47\textwidth]{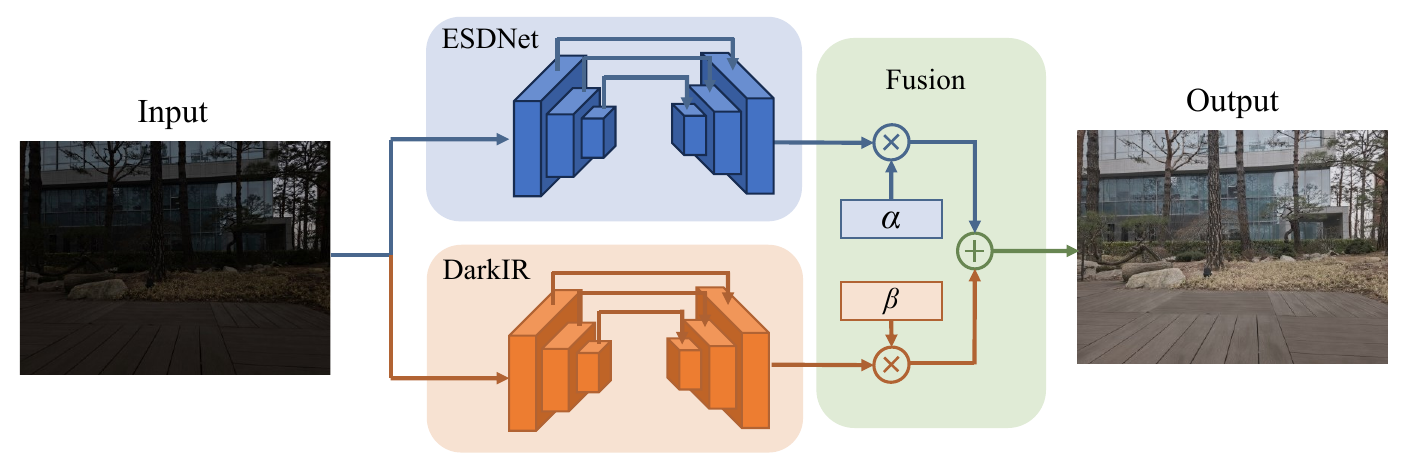}
\end{center}
\vspace{-1em}
\caption{Pipeline of RLLIE proposed by \emph{SYSU-FVL}}
\label{fig:RUHDLLIE}
\vspace{-1em}
\end{figure}

\subsection*{WHU-MVP}

\paragraph{Training Details.} They utilize the full-resolution images from the LSD dataset provided by the organizers, rather than the cropped patch version. Training is performed exclusively on four NVIDIA RTX 4090 GPUs. A progressive training strategy is adopted, where the patch sizes are set to ${960, 1280, 1280, 1440, 1600}$, with corresponding batch sizes of ${4, 2, 2, 2, 1}$, and training iterations of ${36\text{K}, 24\text{K}, 12\text{K}, 28\text{K}, 50\text{K}}$, respectively. The initial learning rate is set to $3 \times 10^{-4}$ scheduled using cyclic cosine annealing.

\subsection*{KLETech-CEVI}

\begin{figure*}[!t]
\centering
\includegraphics[width=\linewidth]{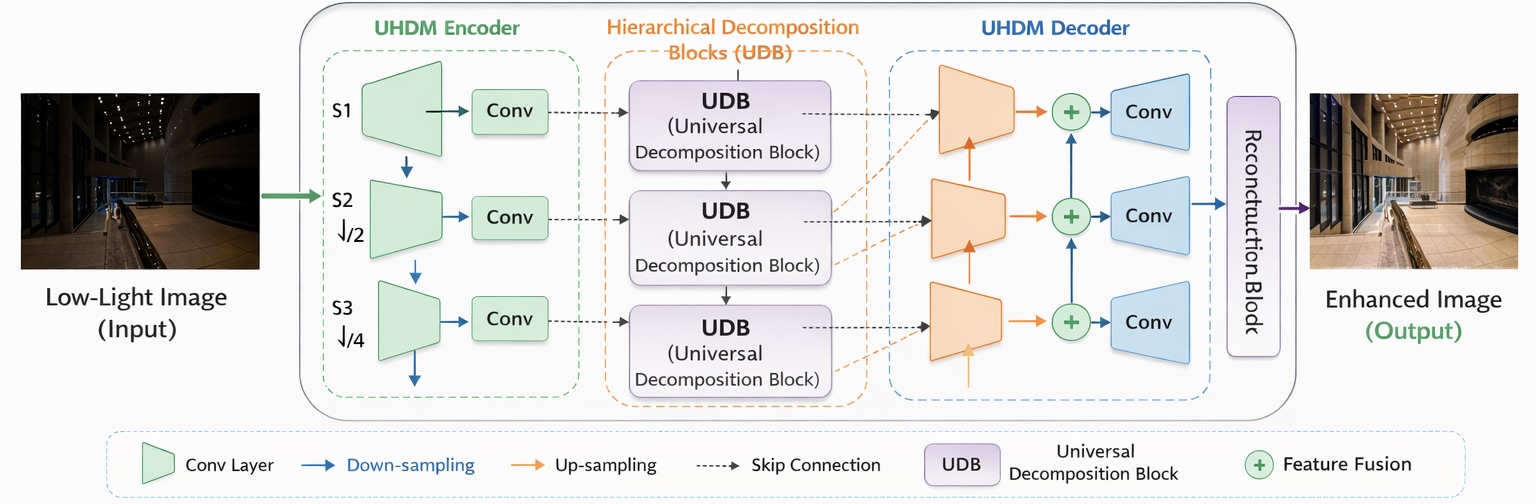}
\vspace{-6pt}
\caption{Overview of the proposed UHDM architecture by \emph{KLETech-CEVI}}
\vspace{-10pt}
\label{fig:team03-kletech-cevi}
\end{figure*}

\paragraph{Training Details} The proposed model is trained using Python and PyTorch Framework with settings as mention in Table \ref{table:kletech-cevi}

\begin{table}[!hbtp]
\centering
\caption{Technical summary of UHDM model proposed by Team KLETech-CEVI.}
\vspace{-2pt}
\small
\setlength{\tabcolsep}{6pt}
\renewcommand{\arraystretch}{1.2}
\resizebox{\columnwidth}{!}{% % Added resizebox here
\begin{tabular}{lcccccccc}
\hline
Input & Time & Iter. & Extra & Diff. & Attn. & Quant. & Params (M) & Runtime \\
\hline
384×384 & 27h & 290k & No & No & Yes & No & 41.5 & 1.2 s \\
\hline
\end{tabular}
} % End of resizebox
\vspace{-6pt}\label{table:kletech-cevi}
\end{table}

\subsection*{BITssvgg}

\paragraph{Training Details.}

% \textcolor{red}{
The total complexity of their method across all stages is 16.87M parameters, with an average runtime of 3.2 seconds on an NVIDIA RTX 4090 48GB GPU.
The model is trained in an end-to-end supervised manner using paired low-light/normal-light image pairs. During training, random patches of size $128 \times 128$ are cropped from the training images as network inputs, and standard geometric data augmentations are applied to improve the robustness and generalization ability of the model. They train the network for a total of 300,000 iterations using the AdamW optimizer. The initial learning rate is set to $3 \times 10^{-4}$, the weight decay is $1 \times 10^{-4}$, and the momentum coefficients are $(0.9, 0.999)$. In addition, gradient clipping is employed during optimization to stabilize training and alleviate potential gradient explosion. For learning rate scheduling, they adopt a two-stage cosine annealing restart strategy. Specifically, the first 92,000 iterations maintain a relatively high learning rate regime for sufficient optimization, while the remaining 208,000 iterations gradually decay the learning rate from $3 \times 10^{-4}$ to $1 \times 10^{-6}$, enabling the model to achieve more stable convergence in the later stage. The overall training objective is mainly based on the $\mathcal{L}_1$ pixel-wise reconstruction loss, which encourages the restored output to be consistent with the 
corresponding normal-light reference image.
% }

% \textcolor{red}{
During testing, they first load the trained DERNet weights and set the model to evaluation mode for image-by-image inference on the GPU. For each input image, they first read and normalize it to the range of $[0, 1]$. Then, an adaptive resizing strategy is applied according to the short side of the input image: if the short side is greater than or equal to $3k$ pixels, the image is resized to $0.5\times$ its original size; if the short side is greater than $2k$ pixels, the image is resized to $0.8\times$ its original size; otherwise, the original resolution is retained. To satisfy the network requirement, the resized image is padded so that both its height and width are multiples of 8. The padded image is then fed into the network to obtain the restored result, after which the padded region is removed. Finally, if resizing was applied before inference, the restored output is resized back to the original resolution. The result is then clipped to the range of $[0, 1]$ and saved as the final output image.
% }

\begin{figure}
\centering
\includegraphics[width=\linewidth]{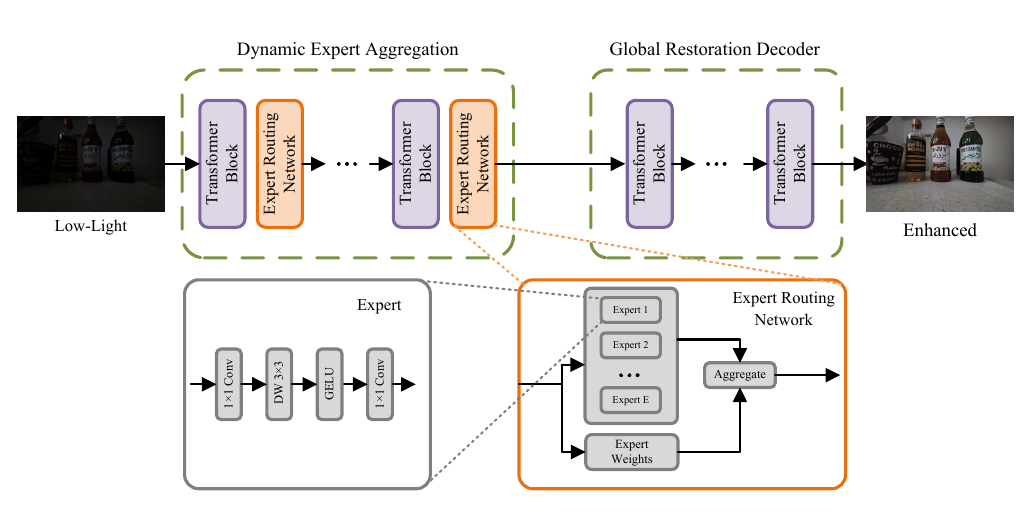} 
\caption{\emph{BITssvgg}'s architecture overview}
\label{fig:bitssvgg}
\end{figure}

\subsection*{BAU-Vision}

\begin{figure*}[htbp]
\begin{center}
\includegraphics[width=\linewidth]{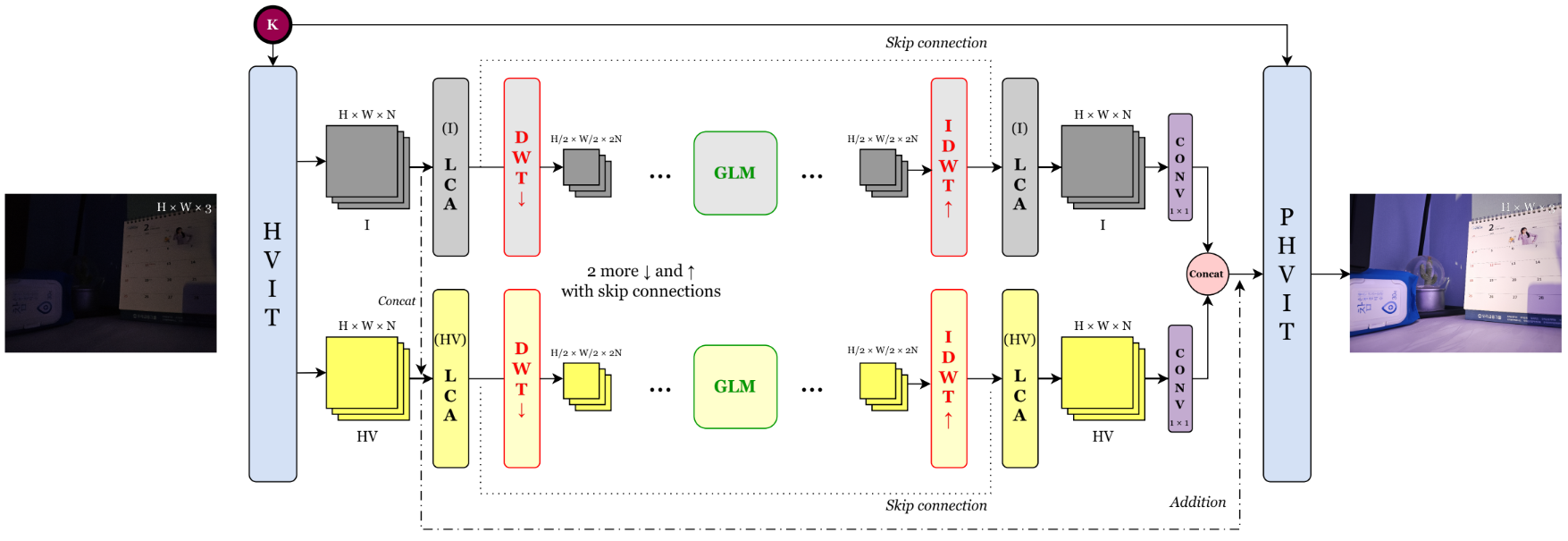}
\end{center}
\vspace{-1em}
\caption{\emph{BAU-Vision}'s Wave-P architecture}
\label{fig:bau-arch}
\vspace{-1em}
\end{figure*}

\paragraph{Training Details.} To stabilize training on given challenge dataset, They define a sample-adaptive coefficient vector $\alpha \in \mathbb{R}^N$ for each batch of $N$ samples. For the $i$-th sample, the coefficient $\alpha_i$ is computed as the maximum ratio between the ground-truth and predicted global mean intensities ($\mu$): $\alpha_i = \max(\mu_{\text{GT}}^i / \mu_{\text{pred}}^i, \mu_{\text{pred}}^i / \mu_{\text{GT}}^i)$. This formulation balances gradient contributions across the batch regardless of absolute brightness levels, thus preventing optimization bias toward high-intensity samples. The network is supervised by a dual-domain multi-objective loss that combines Charbonnier \cite{charbonnier1997deterministic}, SSIM, and Laplacian edge terms with \textit{VGG-19} based perceptual similarity \cite{johnson2016perceptual} to enforce joint pixel-level and structural consistency. The optimization is performed using the AdamW \cite{loshchilov2017decoupled} regulated by cosine annealing scheduler. To enhance model robustness against diverse lighting conditions, they incorporate random gamma adjustments ($\gamma \in [0.6, 1.2]$) during training and employ 6-view TTA during the inference.

\subsection*{SNUCV}

\paragraph{Training Details.}
They trained CIDNet+~\cite{yan2025hvi} and OSEDiff~\cite{wu2024one} exclusively on the LSD training dataset~\cite{sharif2026illuminating}. For CIDNet+, they followed the original paper's protocol and cropped patches to a size of $1280 \times 1280$. OSEDiff was initialized from a pre-trained Stable Diffusion checkpoint (CompVis/stable-diffusion-v1-4) and fine-tuned on the LSD dataset according to the ICM-SR~\cite{kang2025icm} training procedure. For ORNet~\cite{ryou2025beyond}, they directly utilized the publicly released pre-trained weights without any additional fine-tuning.

\begin{figure}[t]
\centering
\includegraphics[width=0.45\textwidth]{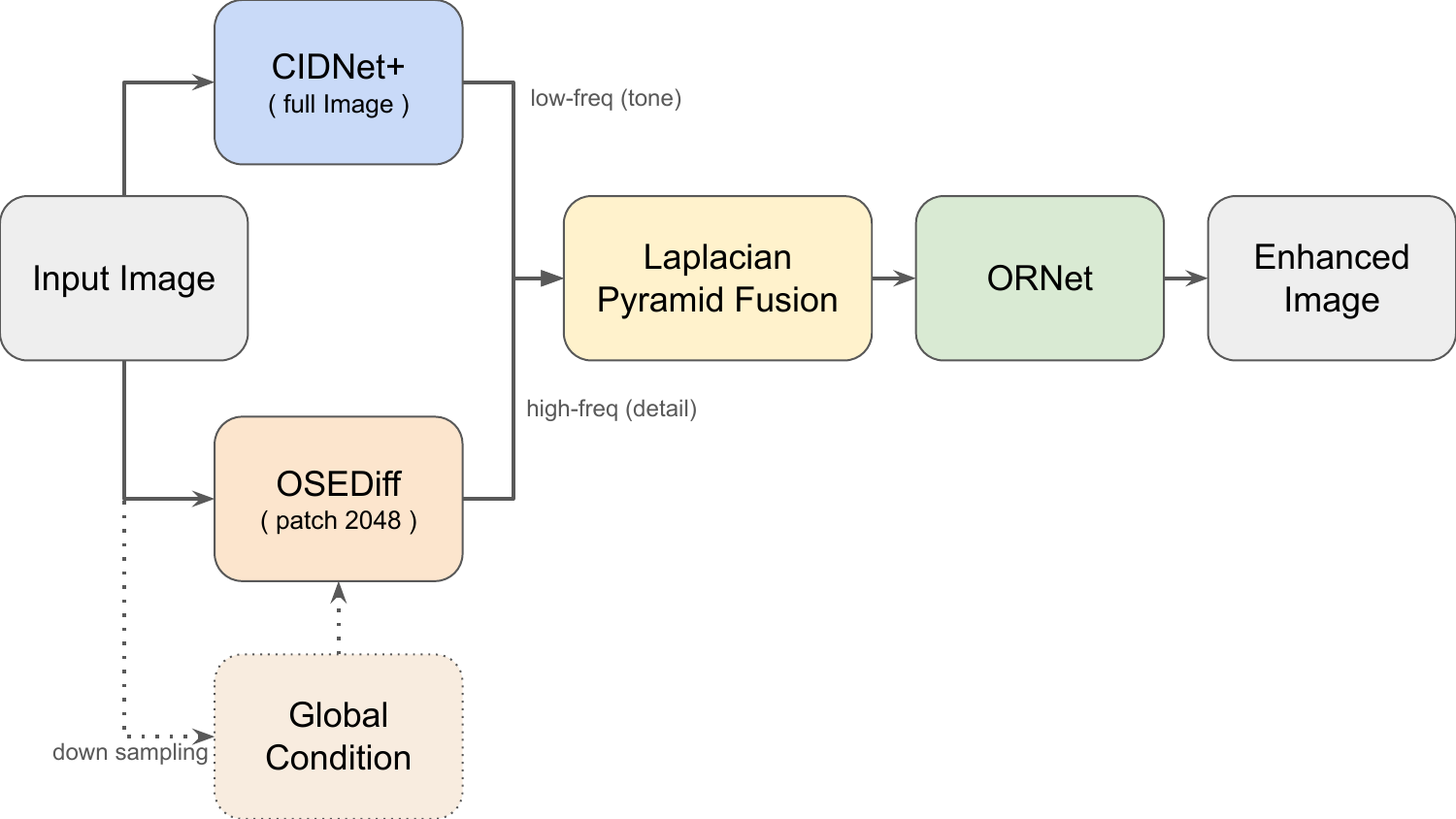}
\caption{\emph{SNUCV}'s MB-LPFR pipeline}
\label{fig:MB-LPFR}
\end{figure}

\subsection*{YuFans}

\begin{figure}[t]
\centering
\resizebox{\columnwidth}{!}{%
\begin{tikzpicture}[
    node distance=0.35cm and 0.4cm,
    block/.style={rectangle, draw, fill=blue!8, text width=1.8cm, minimum height=0.65cm, align=center, rounded corners=2pt, font=\scriptsize},
    stage/.style={rectangle, draw=orange!70, fill=orange!10, text width=1.8cm, minimum height=0.65cm, align=center, rounded corners=2pt, font=\scriptsize\bfseries},
    data/.style={rectangle, draw=green!60!black, fill=green!8, text width=1.6cm, minimum height=0.55cm, align=center, rounded corners=2pt, font=\scriptsize},
    novel/.style={rectangle, draw=red!70, fill=red!8, text width=1.8cm, minimum height=0.65cm, align=center, rounded corners=2pt, font=\scriptsize\bfseries},
    arrow/.style={-{Stealth[length=2mm]}, thick},
]
% Stage 1
\node[data] (lsd) {LSD Patches\\115K, 512$^2$};
\node[stage, right=of lsd] (s1) {Pretrain\\5 epochs};
\node[block, right=of s1] (bb) {NAFNet w64\\116M};

% Stage 2
\node[data, below=0.7cm of lsd] (msdb) {\textbf{MSDB}\\200 pairs, 4K};
\node[novel, right=of msdb] (baa) {\textbf{BAA}\\$\gamma\!\in\![0.7,2.5]$};
\node[stage, right=of baa] (s2) {\textbf{PPE}\\384$\to$768};

% Inference
\node[block, below=0.7cm of s2] (tile) {Tile 768\\+8$\times$TTA};
\node[block, left=of tile] (out) {Enhanced\\4K Output};

% Arrows
\draw[arrow] (lsd) -- (s1);
\draw[arrow] (s1) -- (bb);
\draw[arrow] (bb) -- node[right, font=\tiny\itshape, text=gray] {weights} (s2);
\draw[arrow] (msdb) -- (baa);
\draw[arrow] (baa) -- (s2);
\draw[arrow] (s2) -- (tile);
\draw[arrow] (tile) -- (out);

% Labels
\node[above=0.01cm of baa, font=\tiny\bfseries, text=red!70] {\textsc{Novel}};
\node[above=0.01cm of msdb, font=\tiny\bfseries, text=red!70] {\textsc{Novel}};
\node[above=0.01cm of s2, font=\tiny\bfseries, text=red!70] {\textsc{Novel}};

% Background
\begin{scope}[on background layer]
\node[draw=blue!20, fill=blue!3, rounded corners=3pt, fit=(lsd)(s1)(bb), inner sep=4pt, label={[font=\tiny\bfseries, text=blue!50]above:Stage 1}] {};
\node[draw=orange!30, fill=orange!3, rounded corners=3pt, fit=(msdb)(baa)(s2), inner sep=4pt, label={[font=\tiny\bfseries, text=orange!60]above:Stage 2}] {};
\end{scope}
\end{tikzpicture}%
}
\caption{BAPE pipeline. \emph{YuFans}'s novel contributions---Brightness-Aware Augmentation (BAA), Multi-Source Domain Bridging (MSDB), and Progressive Patch Escalation (PPE).}
\label{fig:bape_pipeline}
\end{figure}
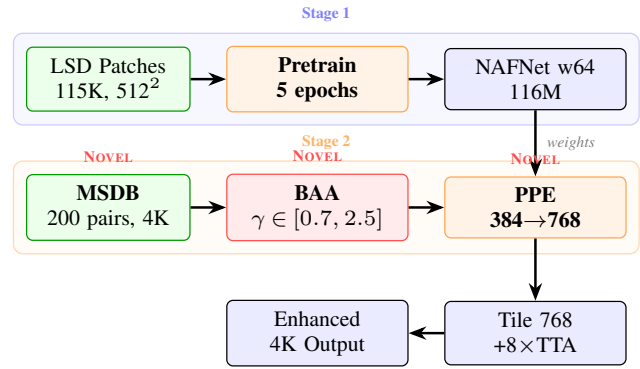

% \textbf{(1) Brightness-Aware Augmentation (BAA):} They apply stochastic gamma correction ($\gamma \sim \mathcal{U}[0.7, 2.5]$) and brightness jittering \textit{exclusively to input images} during training, simulating extreme low-light conditions without modifying ground truth targets.

% \textbf{(2) Progressive Patch Escalation (PPE):} A multi-stage fine-tuning schedule ($384 \rightarrow 512 \rightarrow 768$) progressively increases spatial context, reducing the patch-to-full-image PSNR gap from 3.7\,dB to 0.75\,dB.

% \textbf{(3) Multi-Source Domain Bridging (MSDB):} The team curates 200 full-resolution pairs from four LSD In-the-wild subsets (DLI, DLO, DEI, DEO), spanning brightness levels 21--42. The DEI/DEO subsets, closest to competition brightness, serve as domain bridges, reducing overfitting vs.\ using only 24 validation pairs.

\paragraph{Training Details.}
Training proceeds in two stages (Table~\ref{tab:bape_schedule}). \textbf{Stage~1:} NAFNet w64 pretrained on 115,376 LSD DLL patches~\cite{sharif2026illuminating} for 5 epochs. \textbf{Stage~2:} Progressive fine-tuning on 200 full-resolution pairs with BAA. Loss: Charbonnier ($\epsilon\!=\!10^{-3}$) + $0.1\!\times\!$SSIM. FP32 with gradient clipping (max norm 0.5) and NaN skip, as AMP caused instability. Total: ${\sim}$9\,h on 2$\times$ H800 80\,GB.

\begin{table}[t]
\centering
\caption{\emph{YuFans}'s Ablation study. Indep.\ Val uses 10 held-out pairs.}
\label{tab:bape_ablation}
\setlength{\tabcolsep}{3pt}
\resizebox{\columnwidth}{!}{%
\begin{tabular}{lcc}
\toprule
\textbf{Configuration} & \textbf{Indep.\ Val} & \textbf{Comp.\ Val} \\
\midrule
NAFNet w96, 24-val only, p384 & --- & 19.02 \\
\quad + LSD pretraining (3\,ep) & --- & 23.53 \\
\quad + Progressive $\to$p768 & --- & 24.13 \\
\quad + MSDB (200 pairs) & 24.54 & 23.79 \\
\textbf{Full BAPE} & \textbf{24.54} & \textbf{23.79} \\
\bottomrule
\end{tabular}%
}
\end{table}

\begin{table}[t]
\centering
\caption{\emph{YuFans}'s Progressive training schedule.}
\label{tab:bape_schedule}
\setlength{\tabcolsep}{3pt}
\resizebox{\columnwidth}{!}{%
\begin{tabular}{lccccc}
\toprule
\textbf{Stage} & \textbf{Data} & \textbf{Patch} & \textbf{Batch} & \textbf{LR} & \textbf{Ep.} \\
\midrule
Pretrain & 115K patches & 256 & 16 & 5e-5 & 5 \\
FT-p384 & 200 full-res & 384 & 8 & 1e-4 & 100 \\
FT-p512 & 200 full-res & 512 & 4 & 5e-5 & 100 \\
FT-p768 & 200 full-res & 768 & 2 & 3e-5 & 100 \\
\bottomrule
\end{tabular}%
}
\end{table}

% During inference, they use tiled processing ($768\!\times\!768$, overlap 128, Hanning window blending) with $8\times$ geometric TTA (4~rotations $\times$ 2~flips). Table~\ref{tab:bape_ablation} shows that LSD pretraining contributes +4.51\,dB, PPE adds +0.60\,dB, and MSDB improves generalization (independent val 24.54\,dB vs.\ competition val 23.79\,dB, a gap of only 0.75\,dB).

During inference, they use tiled processing ($768\!\times\!768$, overlap 128, Hanning window blending) with $8\times$ geometric TTA (4~rotations $\times$ 2~flips). LSD pretraining contributes +4.51\,dB, PPE adds +0.60\,dB, and MSDB improves generalization (independent val 24.54\,dB vs.\ competition val 23.79\,dB, a gap of only 0.75\,dB).

\subsection*{AAIR-ARM}

\paragraph{Training Details.}
They use a frozen Stable Diffusion VAE \cite{stablediff} and an HDiT-XL/4 backbone~\cite{hdit} to parameterize the flow model. The network is trained from scratch on paired low-normal light images from the Smartphone Dataset (LSD)~\cite{sharif2026illuminating}. Training is performed on $2048 \times 2048$ image patches (randomly cropped from the 4K images) with a batch size of 8 for 80K optimization steps using AdamW. The model is optimized with the standard flow-matching objective on the conditional velocity field~\cite{lipman2023flowmatchinggenerativemodeling}, with noise level $\sigma=0.01$. All experiments are conducted using PyTorch on 8 NVIDIA H100 GPUs. At inference, the ODE is solved using 25 function evaluations (NFE).

\begin{figure}[ht]
\centering
\includegraphics[
    width=0.9\linewidth,
    trim=0cm 0cm 4cm 0cm,
    clip
]{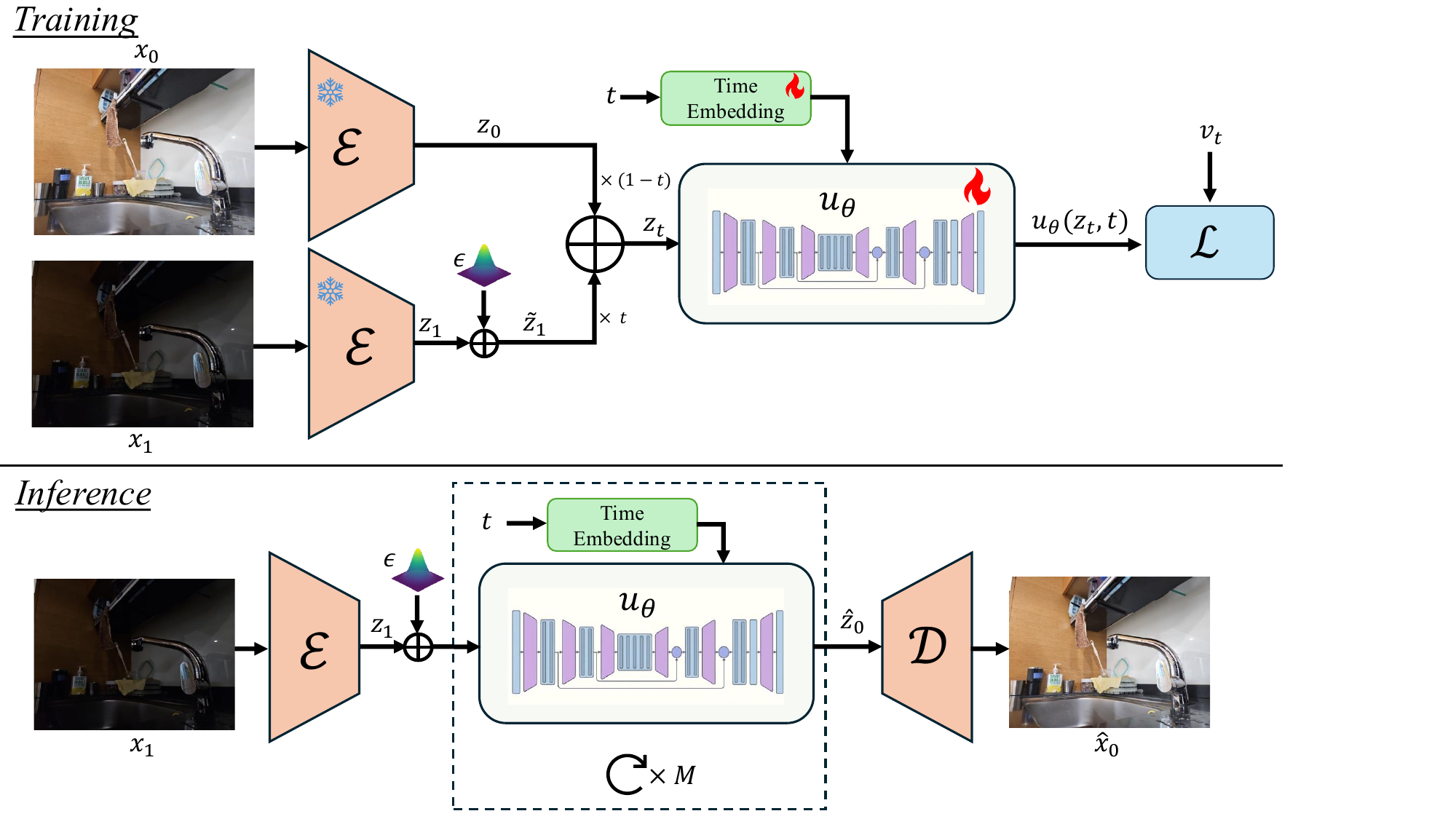}
\caption{\textbf{\emph{AAIR-ARM}'s LFM-LLIE overview.} \textbf{(a) Training:} A flow-matching network $u_\theta$ (HDiT-based) is trained in latent space using interpolated samples $z_t$ created from the noise-perturbed low-light latent image $\tilde{z}_1$. \textbf{(b) Inference:} Starting from $z_1 = \mathcal{E}(x_1)$, the latent is iteratively transformed by the learned velocity field $u_\theta$ into $\hat{z}_0$ and decoded by $\mathcal{D}$ to produce the enhanced image $\hat{x}_0$.}
\label{fig:aairarm}
\end{figure}

\subsection*{TranssionAI}

\begin{figure}[ht]
\centering
\includegraphics[width=0.9\linewidth]{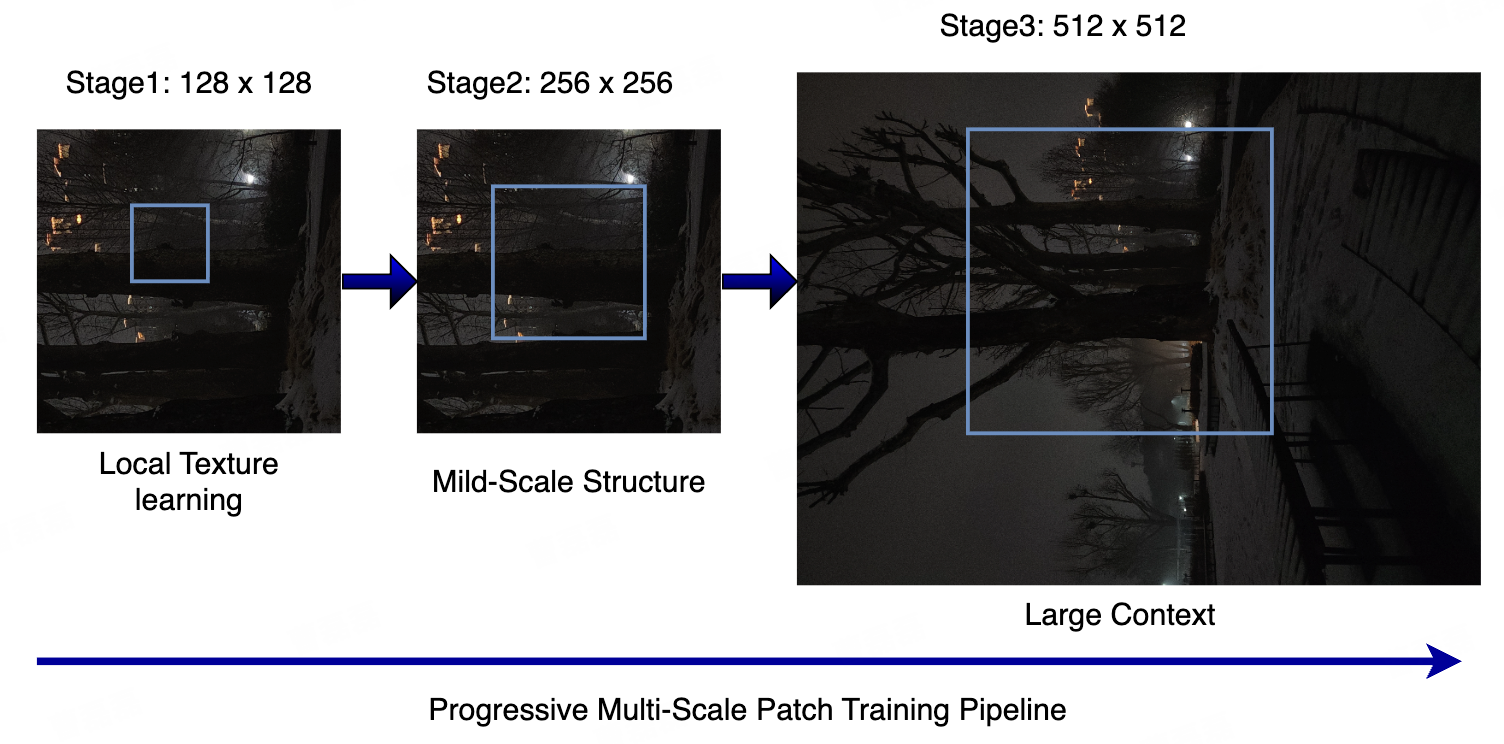}
\caption{\emph{TranssionAI} overview}
\label{fig:transsitionai}
\end{figure}

\paragraph{Training Details.}

% \textcolor{red}{
They optimize the model using a hybrid loss function that combines reconstruction, perceptual, structural, and frequency-domain constraints:
\begin{equation}
L = 0.2L_{r} + 0.2L_{lc} + L_{lpips} + 0.2L_{grad} + 0.5L_{freq}
\end{equation}
where $L_{r}$ is the mean intensity reconstruction loss (GT-Mean loss was adopted, instead of L1 loss), and $L_{lc}$ denotes the luminance and chrominance loss adopted from the baseline method. $L_{lpips}$ improves perceptual quality, while $L_{grad}$ preserves structural details by constraining image gradients. The frequency-domain loss $L_{freq}$ further enhances high-frequency components and fine textures.
All loss weights are empirically determined and remain fixed throughout training. They adopt a progressive multi-scale training strategy, where the model is trained from small to large patch sizes ($128 \to 256 \to 512$), allowing it to gradually capture both local and global image characteristics.
% }

% Testing description: 

% \textcolor{red}{
\paragraph{Testing description.} To process high-resolution images efficiently, they employ a sliding window inference strategy with two window sizes (2048 and 512). The larger window captures global illumination consistency, while the smaller window focuses on recovering fine-grained details. In addition, they apply self-ensemble during inference by performing 8 geometric augmentations, including horizontal and vertical flips as well as rotations. The predictions are inversely transformed and averaged to produce the final result. This combination of multi-scale sliding window inference and self-ensemble enhances both global consistency and local detail reconstruction, leading to improved performance.
% }

% \textcolor{red}{
Quantitative and qualitative advantages of the proposed solution: Shown in Table \ref{tab:quantitative}.
% }

\begin{table*}[ht]
    \centering
    \resizebox{\linewidth}{!}{
    \begin{tabular}{c|c|c|c|c|c|c}
        Stages & PSNR & SSIM & LPIPS & NIQE & MUSIQ & CLIP-IQA  \\
        \hline
        Stage 1 (slide window 1024) & 18.717 & 0.5926 & 0.4427 & 4.851 & 30.1301 & 0.3722 \\
        Stage 2 (slide window 1024)  & 19.078 & 0.5969 & 0.4270 & 4.825 & 30.4576 & 0.3827 \\
        Stage 2 (slide window 1024 + self-ensemble) & 19.166 & 0.6249 & 0.4462 & 4.37 & 31.5657 & 0.4081 \\
        Stage 3 (slide window 2048 + self-ensemble) & 19.9456 & 0.6358 & 0.4592 & 4.689 & 30.1128 & 0.4203 \\
        Stage 3 (multi-scale slide window 2048 and 512 + self-ensemble) & 20.1088 & 0.6328 & 0.4578 & 4.652 & 30.1757 & 0.4146 \\
        
    \end{tabular}
    }
    \caption{Ablation study of \emph{TranssionAI}'s methods on the validation dataset.}
    \label{tab:quantitative}
\end{table*}

\subsection*{ReagvisLabs}

\begin{figure*}[t]
\centering
\includegraphics[width=0.98\textwidth]{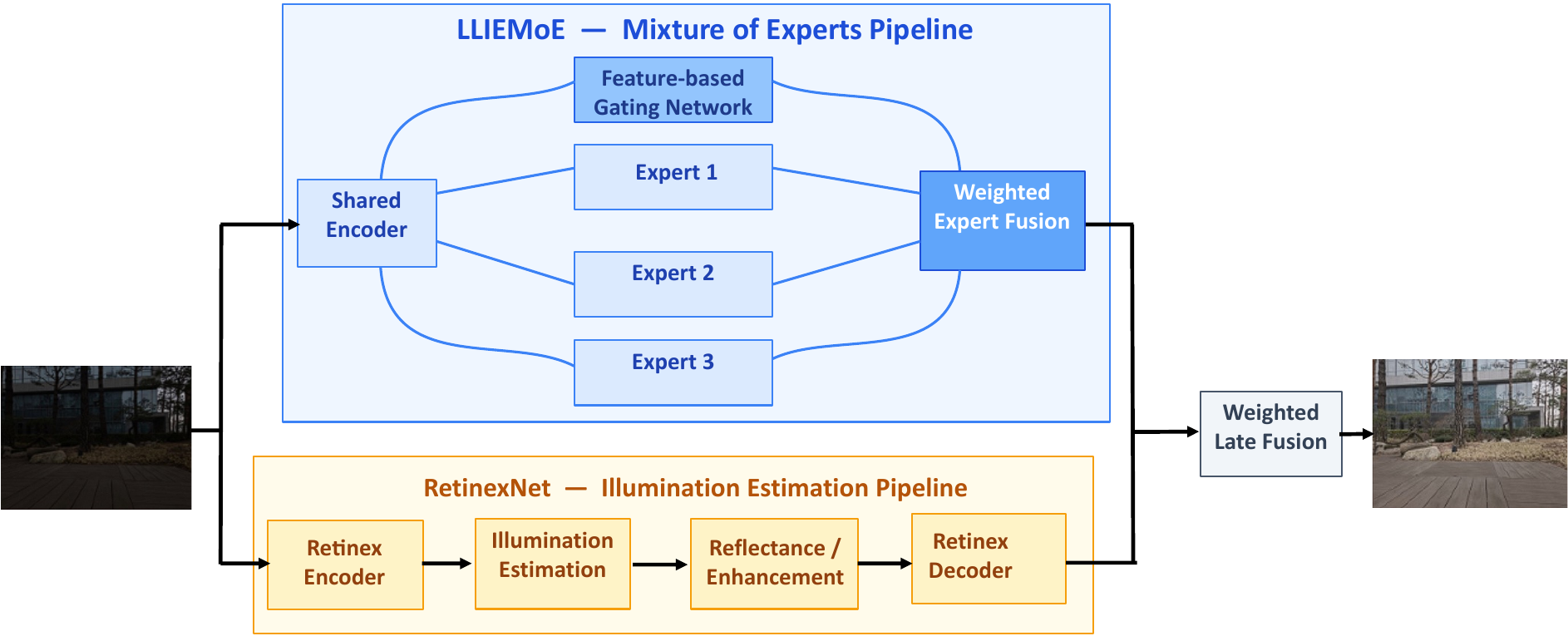}
\caption{
Overview of the proposed weighted late-fusion architecture by \emph{ReagvisLabs}. The input low-light image is processed by two complementary branches. The upper branch, \textbf{LLIEMoE --- Mixture of Experts Pipeline}, first extracts shared encoder features and then uses a feature-based gating network to guide three specialized experts for fidelity-oriented, perceptual, and naturalness-aware restoration, whose responses are aggregated through weighted expert fusion to produce the MoE enhanced output. The lower branch, \textbf{RetinexNet --- Illumination Estimation Pipeline}, performs Retinex encoding, illumination estimation, reflectance/enhancement processing, and decoding to generate an illumination-aware enhanced output. The predictions from both branches are finally combined through weighted late fusion to obtain the final enhanced image.
}
\label{fig:reagvis_architecture}
\end{figure*}

\paragraph{Training Details.}
Both branches are trained in two stages using paired low-light supervision. They first perform mixed pretraining on LSD~\cite{sharif2026illuminating} together with SID~\cite{chen2018sid} to improve generalization across diverse low-light conditions, and then fine-tune on LSD only for domain-focused adaptation. Optimization is performed using AdamW~\cite{loshchilov2019adamw}. The training objective combines pixel-level fidelity, perceptual consistency, structural preservation, and color stability through an $L_1$ loss, VGG perceptual loss~\cite{simonyan2015vgg}, MS-SSIM~\cite{wang2003msssim}, and color consistency regularization. For the LLIEMoE branch, they additionally apply expert specialization supervision and a load balancing regularizer to prevent expert collapse and ensure that different experts remain active during routing. During inference, they use geometric test time augmentation over three scales and average the inversely transformed predictions to obtain the final restored image.

\subsection*{DH-XHDL-Team}

\begin{figure*}
    \centering
    \includegraphics[width=1\linewidth]{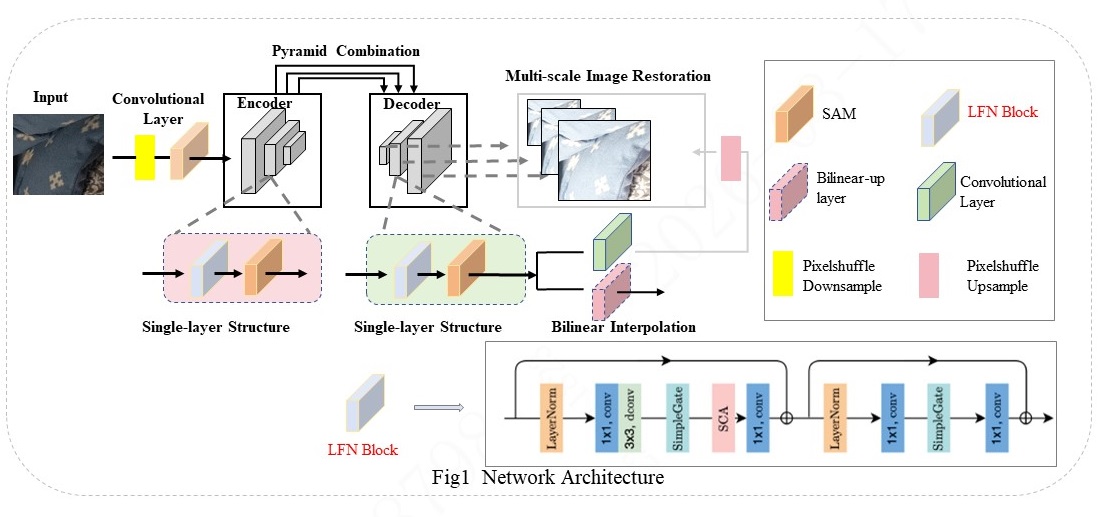}
    \caption{\emph{DH-XHDL-Team}'s TEI-LLIE architechture}
    \label{fig:placeholder}
\end{figure*}

\paragraph{Training Details.}\textbf{(1) Dataset and Augmentation} The training pipeline combines the official NTIRE 2026 LLIE Challenge training/validation data with an additional 110,000 external images for dataset augmentation. No explicit pre-processing operations are applied to the training data. Training samples are constructed via random cropping during the training phase.\\\textbf{(2) Implementation and Optimization} The model is implemented in Python using the PyTorch deep learning framework. Training is conducted on a single NVIDIA L40 GPU, with a total training duration of approximately 112 hours. The Adam optimizer is adopted with an initial learning rate of 0.0001. No additional efficiency optimization strategies are employed beyond the architectural designs of LFNblock.

\begin{table}[t]
\centering
\caption{\emph{DH-XHDL-Team}'s ablative experiments}
\label{tab:TEI-LLIE}
\setlength{\tabcolsep}{3pt}
\resizebox{\columnwidth}{!}{%
\begin{tabular}{lccccc}
\toprule
\textbf{Experiments} & \textbf{LR-cycle} & \textbf{Patch} & \textbf{Batch} & \textbf{LR} & \textbf{Epoch} \\
\midrule
Baseline & 5 & 512 & 16 & 1e-4 & 75 \\
Baseline& 5 & 512 & 16 & 2e-4 & 75 \\
Baseline + LFN & 25 & 512 & 16 & 1e-4 & 50 \\
Baseline + LFN  & 25 & 512 & 8 & 1e-4 & 50 \\
\bottomrule
\end{tabular}%
}
\end{table}

\subsection*{NTR}

\paragraph{Training Details.}
They fine-tune the MDAE-pretrained backbone on the NTIRE 2026 Low Light Enhancement Track~1 training set, consisting of 115{,}376 paired $512 {\times} 512$ patches (low-light input and normal-light ground truth).
The training objective is a composite pixel loss:
\begin{equation}
  \mathcal{L} = \mathcal{L}_{\text{MSE}} + \mathcal{L}_{\text{L1}} + 0.5\,\mathcal{L}_{\text{Sobel}},
\end{equation}
where $\mathcal{L}_{\text{Sobel}}$ is the $\ell_1$ distance between Sobel edge maps of the prediction and ground truth, encouraging preservation of structural detail.
They use AdamW ($\beta_1{=}0.9$, $\beta_2{=}0.999$, weight decay $10^{-3}$) with an initial learning rate of $10^{-4}$ and cosine annealing over 300 scheduled epochs.
An exponential moving average (EMA) of the weights is maintained with a decay of 0.999.
Data augmentation includes random horizontal and vertical flips and $90^{\circ}$ rotations.
Training was conducted on 2$\times$ NVIDIA H200 GPUs (80\,GB each) with a per-GPU batch size of 8.
The best checkpoint was selected at epoch 67 based on validation PSNR on the official 24-image validation set.
No extra data beyond the provided competition training set was used for fine-tuning.

\paragraph{Inference.}
Full-resolution test images ($4080 {\times} 3060$) are processed via tiled sliding-window inference with $512 {\times} 512$ tiles and 256-pixel overlap.
Overlapping regions are blended using a 2D Hann window to suppress tile boundary artifacts.
Input pixel values are linearly mapped from $[0, 1]$ to $[-1, 1]$; the output is mapped back to $[0, 1]$ before saving.
No test-time augmentation or model ensembling is applied.
Inference takes approximately 9 seconds per full-resolution image on a single NVIDIA A10 (24\,GB).

\subsection*{VesperLux}

\paragraph{Training Details.}
The model is trained for $300\text{k}$ iterations using the Adam optimizer ($\beta_1{=}0.9$, $\beta_2{=}0.999$) with an initial learning rate of $2{\times}10^{-4}$ and gradient clipping. The learning rate schedule consists of two phases: a constant phase for the first $92\text{k}$ iterations at $3{\times}10^{-4}$, followed by cosine annealing over the remaining $208\text{k}$ iterations from $3{\times}10^{-4}$ to $10^{-6}$. Data augmentation includes MixUp with $\beta{=}1.2$, retaining the identity sample with a fixed probability.

\subsection*{PSU}

\begin{figure*}[t]
  \centering
  \includegraphics[width=\textwidth]{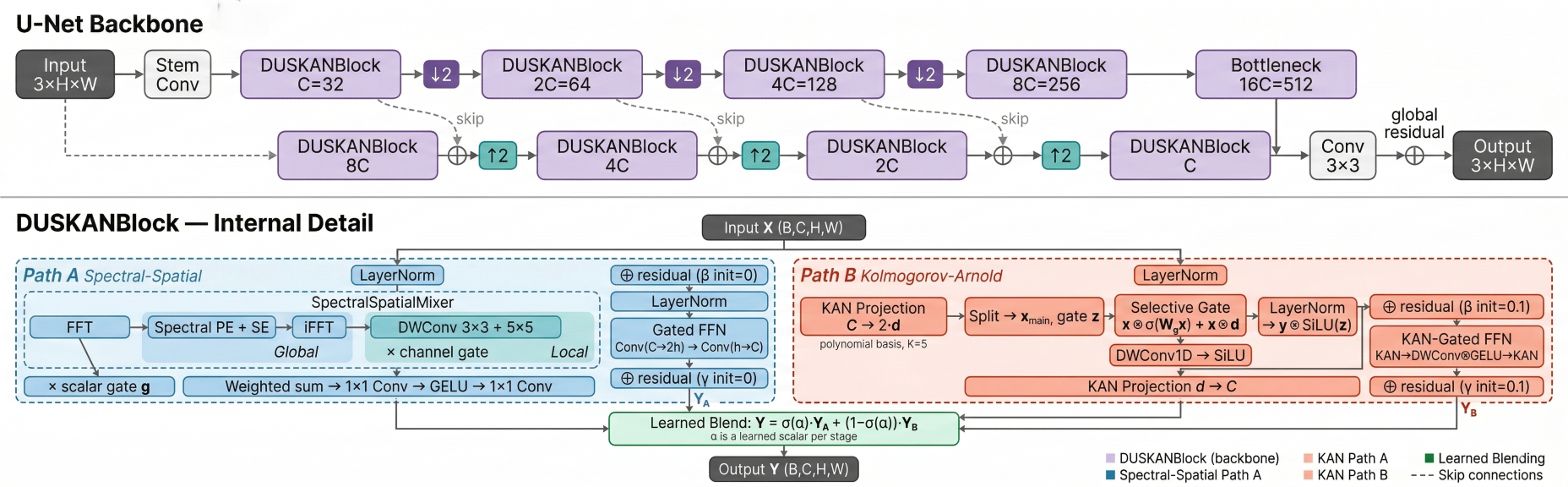}
  \caption{\textbf{DUSKAN architecture by \emph{PSU}.} \textit{Top}: Symmetric 4-level U-Net with DUSKANBlock stages, strided downsampling, PixelShuffle upsampling, and global residual learning. \textit{Bottom}: DUSKANBlock detail. \textbf{Path A} (blue) extracts global features via FFT magnitude modulation and fuses them with local multi-scale depthwise convolution features. \textbf{Path B} (red) uses Kolmogorov-Arnold polynomial-basis activations with a parallel-additive selective gate.}
  \label{fig:duskan_arch}
\end{figure*}

\paragraph{Training Details.} 
The network (45.6M parameters) was trained from scratch for 500 epochs with a batch size of 2 on a single NVIDIA A100 (80GB) GPU. Training utilized $512 \times 512$ random patches with geometric augmentations (random rotations and flips). They employed the AdamW optimizer ($\beta_1 = 0.9, \beta_2 = 0.999$, weight decay $= 10^{-4}$) with an initial learning rate of $2 \times 10^{-4}$, gradually decreased to $1 \times 10^{-6}$ via a Cosine Annealing scheduler. PyTorch's Automatic Mixed Precision (AMP) with FP16 gradients was used to maximize efficiency, resulting in a training time of roughly 49 hours. During inference, an overlapping patch-based strategy was applied to reconstruct the full-resolution test images.

\subsection*{SOMIS-LAB}

\paragraph{Training Details.}

To ensure stable training on the provided challenge dataset, They crop the original images into $512\times512$ patches. The network is optimized using the AdamW optimizer with an initial learning rate of $ 3e-4$  for 200 epochs. To enhance the generalization capability of the model, data augmentation techniques, including random horizontal flipping and random rotations, are applied during the training phase. All experiments are implemented using the PyTorch framework and conducted on an NVIDIA RTX 3090 GPU. Their comprehensive loss function comprises an L1 loss, a perceptual loss, and an SSIM loss, ensuring that the enhanced images maintain high fidelity with the ground truth references at pixel, perceptual, and structural levels.

\subsection*{MC2}

\begin{figure}[htbp]
\begin{center}
\includegraphics[width=0.47\textwidth]{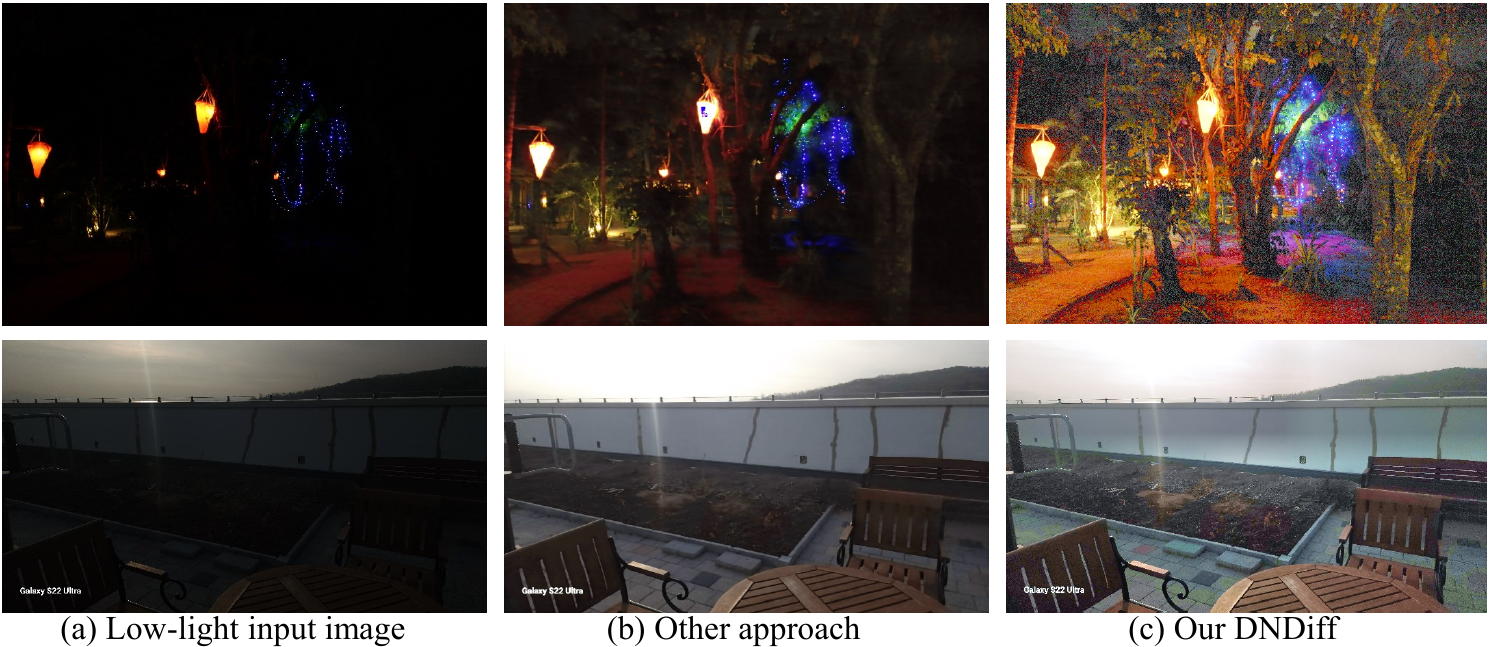}
\end{center}
\vspace{-1em}
\caption{Qualitative comparison of DNDiff's color performance for low-light enhancement proposed by \emph{MC2}.}
\label{fig:DNDiff}
\vspace{-1em}
\end{figure}

\paragraph{Training Details.}
Notably, the DNDiff is a training-free approach, which bypasses the need for extensive retraining on specific noise distributions.

\subsection*{AnanyaHariniLaksh}

\paragraph{Training Details.}

% \textcolor{red}{
They used the training data provided by the challenge organizers. Training
time - approximately 30 minutes. They performed fine-tuning on a pre-trained model.
% }

\subsection*{Lucky-one}

\begin{figure*}[htbp]
\begin{center}
\includegraphics[width=\textwidth]{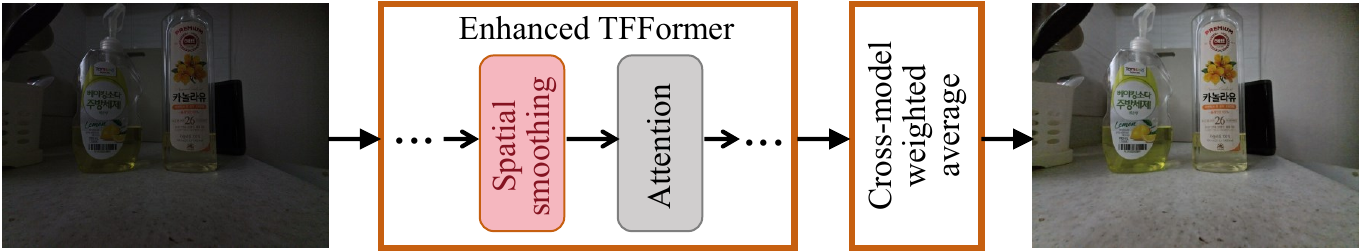}
\end{center}
\vspace{-1em}
\caption{DNATT pipeline of Team Lucky one.}
\label{fig:DNATT}
\vspace{-1em}
\end{figure*}

\paragraph{Training Details.}
They introduce customized data augmentation strategies to retrain the improved network, employing a learning rate of 1e-4 to ensure stable and effective convergence. Trained on an NVIDIA GeForce RTX 4090 GPU. Furthermore, they re-calibrate the loss function weights to prioritize the restoration of Peak Signal-to-Noise Ratio (PSNR). The resulting composite loss function is defined as:
\begin{equation}
  \mathcal{L} = \mathcal{L}_{\text{MAE}} + \mathcal{L}_{\text{Luminance}},
\end{equation}
Additionally, an ensemble strategy employing a cross-model weighted average is incorporated during the inference phase, significantly enhancing the model's overall robustness and predictive accuracy.

\subsection*{APRIL-AIGC}

\begin{figure}[t]
\centering
\includegraphics[width=\columnwidth]{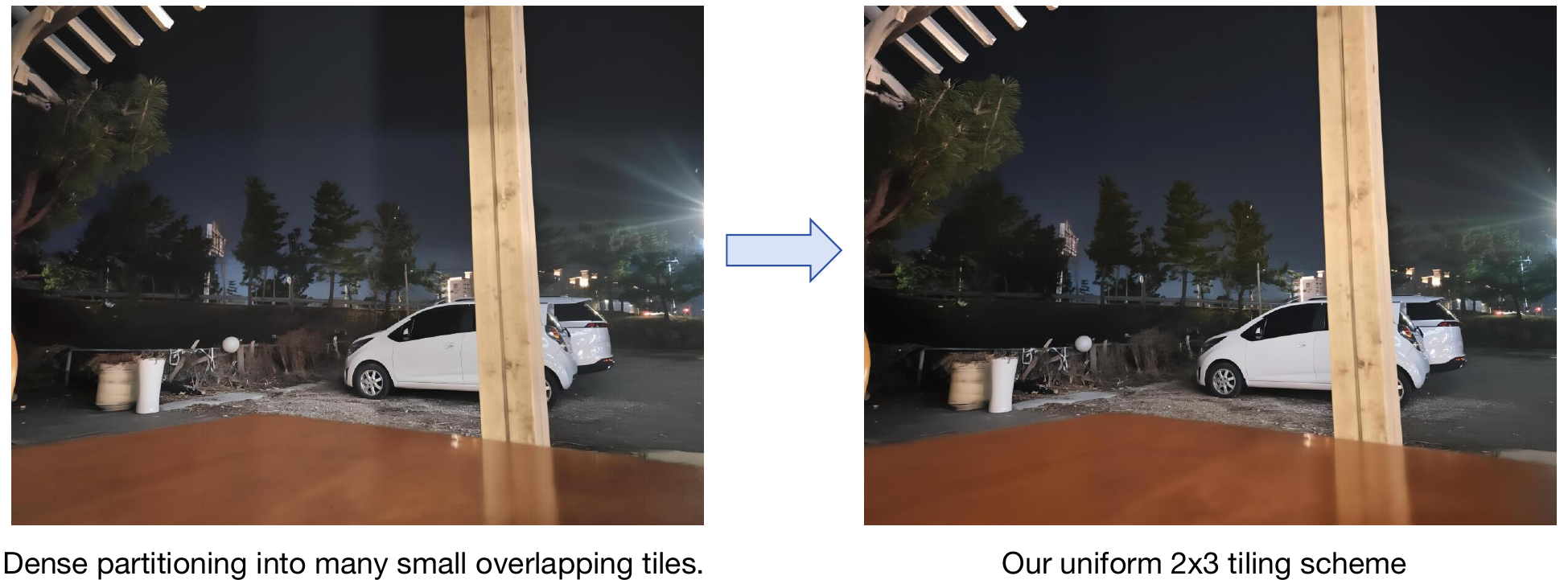}
\caption{\emph{APRIL-AIGC}'s comparison of tiling strategies for high-resolution inference. A fixed $2 \times 3$ partition with moderate overlap yields fewer illumination discontinuities than denser window layouts.}
\label{fig:team19_tiling}
\end{figure}

\paragraph{Training Details.}
Training follows the standard FLUX.2 image-conditioning setting. They fine-tune the model on the released training pairs using random $512 \times 512$ crops for 10,000 iterations on 8 $\times$ NVIDIA H20 GPUs. The optimizer is AdamW with a learning rate of $1 \times 10^{-5}$, a global batch size of 64, and a text-drop probability of 0.15. No extra training data is used. They keep a fixed prompt that emphasizes denoising, shadow recovery, natural brightness adjustment, and color fidelity.

The diffusion backbone is optimized with the rectified-flow objective~\cite{liu2022flow}. Given a clean latent $z_0$ and Gaussian noise $\epsilon \sim \mathcal{N}(0, I)$, the forward interpolation is
\[
z_t = (1 - t) z_0 + t \epsilon,
\]
with target velocity
\[
u_t = \epsilon - z_0.
\]
The network prediction $v_{\theta}(z_t, t)$ is trained by
\[
L_{\mathrm{RF}} = \|v_{\theta}(z_t, t) - u_t\|_2^2.
\]
At inference time, each full-resolution image is decomposed into six overlapping tiles arranged in a fixed $2 \times 3$ grid. For a $3060 \times 4080$ image, each tile is approximately $1600 \times 1456$ pixels with overlaps close to $128 \times 144$ pixels. They use 5 sampling steps, a guidance scale of 2.0, and an empty string negative prompt. In practice, this setting provides the most stable trade-off between local restoration quality and global illumination consistency.

\subsection*{MiVideoDLLIE}

\begin{figure}[ht]
\begin{center}
\includegraphics[width=0.47\textwidth]{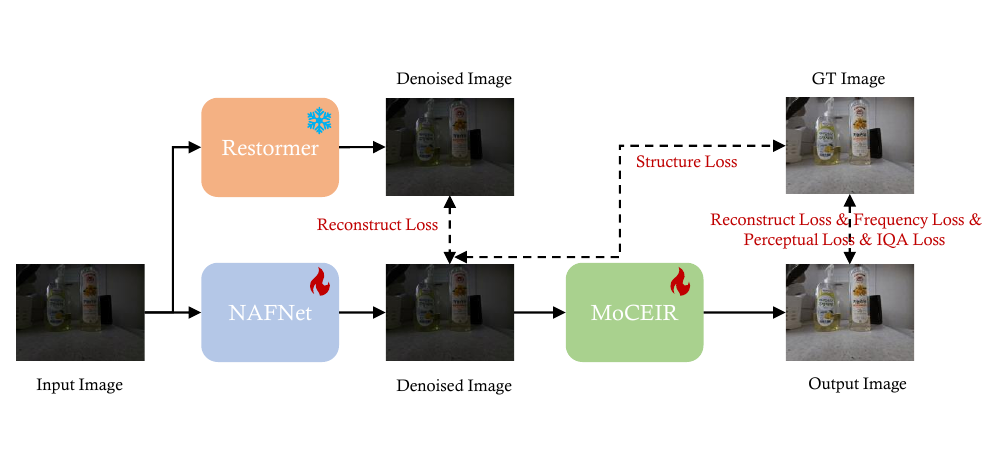}
\end{center}
\vspace{-1em}
\caption{\emph{MiVideoDLLIE}'s pipeline of MiDLLIE}
\label{fig:MiDLLIE}
\vspace{-1em}
\end{figure}

\paragraph{Training Details.}

During the training phase, to improve training efficiency while preserving image high-frequency details as much as possible, they resize the training data from $512\times512$ to $256\times256$. The optimizer is Adam with hyperparameters $\beta_1 = 0.9$ and $\beta_2 = 0.99$, and the initial learning rate is set to $1 \times 10^{-4}$, using a cosine annealing with restarts strategy. For the denoising stage, the NAFNet output is supervised by both the Restormer prediction and the ground truth via reconstruction loss and structural loss, with loss weights of 1 and 0.08, respectively. For the low-light enhancement stage, the MoCEIR output is optimized using reconstruction loss, frequency-domain loss, perceptual loss, and IQA loss against the ground truth, with respective weights of 1, 0.01, 0.02, and 0.02.

\subsection*{RetinexDualV2}

\paragraph{Training Details}
The approach was implemented using BasicSR with PyTorch, the network was trained from scratch for 224K iterations ($\sim$45 epochs) on a single NVIDIA H100 GPU with 4.8M parameters. Training strictly utilized the provided 64,318 image pairs from the NTIRE 2026 JNLLIE. A batch size of 13 and $512 \times 512$ patches were used. They employed the AdamW optimizer ($\beta_1 = 0.9, \beta_2 = 0.999$, weight decay $= 0.001$) with an initial learning rate of $2 \times 10^{-5}$ gradually decreased to $1 \times 10^{-7}$, aided by gradient clipping and PyTorch's Automatic Mixed Precision (AMP) to maximize efficiency ($\sim$133 hours training time). Following a deep supervision strategy, they minimized a robust multi-scale objective:
$\mathcal{L}_{total} = \sum_{i=1}^{3} w_i \big( \mathcal{L}_{cb} + \lambda_{ssim}(1 - \text{SSIM}) + \lambda_{fft}\mathcal{L}_{fft} + \lambda_p \mathcal{L}_p \big)$
merging Charbonnier \cite{DBLP:journals/corr/abs-1710-01992}, SSIM \cite{1284395}, Fast Fourier Transform (FFT), and AlexNet perceptual loss \cite{NIPS2012_c399862d, 8578166} progressively scaled over three feature scales ($w_i = [0.25, 0.5, 1.0]$). During inference, full-resolution test images were padded reflectively to multiples of 128 and processed in a single forward pass without sliding-window cropping (avg. inference runtime of 2.97s).

\begin{figure*}
\centering
\includegraphics[width=\linewidth]{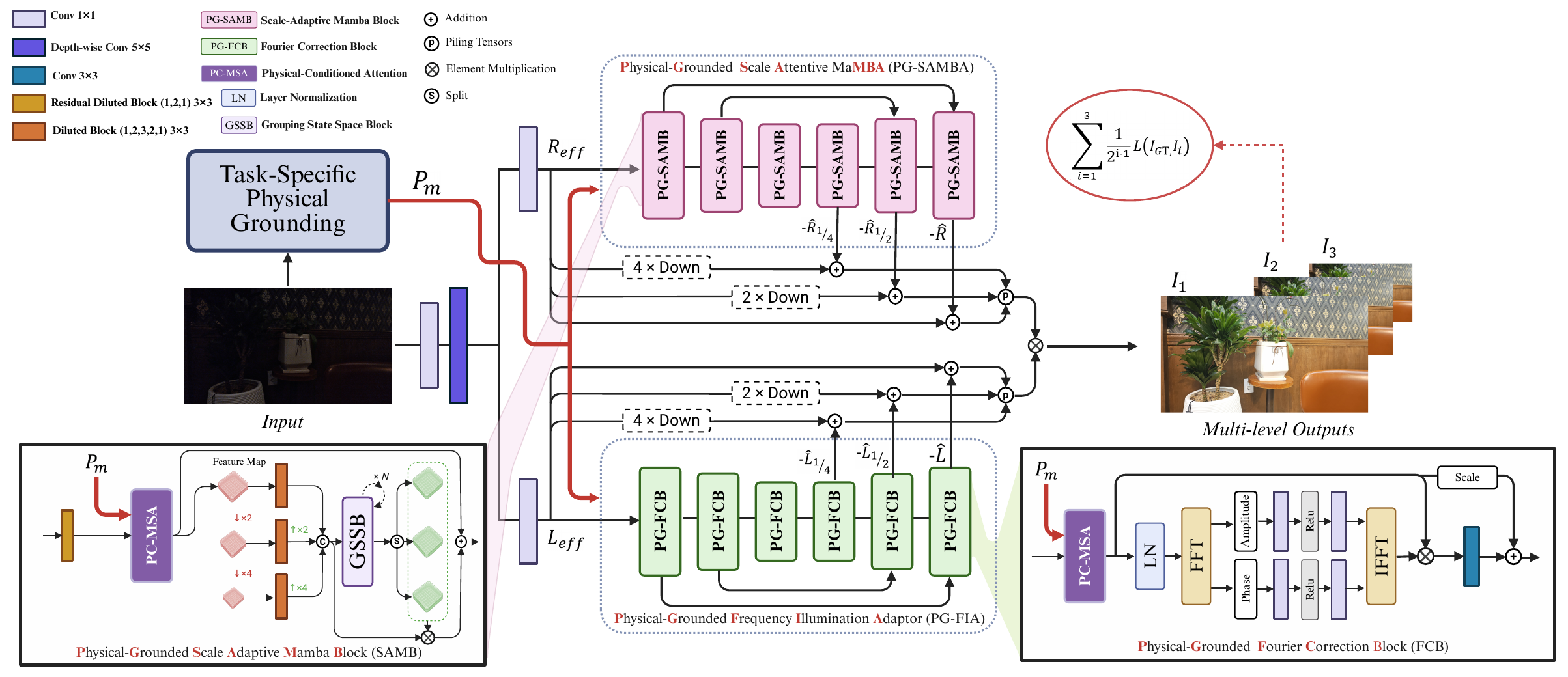} 
\caption{RetinexDualV2 overview}
\label{fig:RD_overview}
\end{figure*}

\subsection*{WIRNet}

\paragraph{Training Details}
They trained the proposed WIRNet with the AdamW
optimizer \cite{loshchilov2017decoupled} on a single NVIDIA RTX 4090 using the LSD NLL
dataset \cite{sharif2026illuminating}. They set the batch size to 16 and the patch size to
$3 \times 256 \times 256$. The total loss function is a weighted combination
of losses computed in both RGB and HVI color spaces:
$\mathcal{L}_{\text{total}} = \mathcal{L}_{\text{RGB}} + \lambda\, \mathcal{L}_{\text{HVI}}$,
where each domain loss is defined as:
$\mathcal{L} = \alpha\, \mathcal{L}_{1} + \beta\, \mathcal{L}_{\text{SSIM}}
+ \gamma\, \mathcal{L}_{\text{Edge}} + \delta\, \mathcal{L}_{\text{LPIPS}}$. They set them to
$\lambda = 1$, $\alpha = 1$, $\beta = 0.5$, $\gamma = 50$, and $\delta = 0.01$ to balance different losses during training.

Fig.~\ref{fig:WIRNet} shows their proposed Wavelet Intensity Refinement Network (WIRNet), which is built upon the CIDNet \cite{yan2025hvi} baseline. While CIDNet achieves competitive performance, it remains sensitive to noise and has limited ability to recover fine details. To address these limitations, they introduce three key contributions: 
\begin{figure}[t]
\centering
\includegraphics[
    width=0.6\linewidth
]{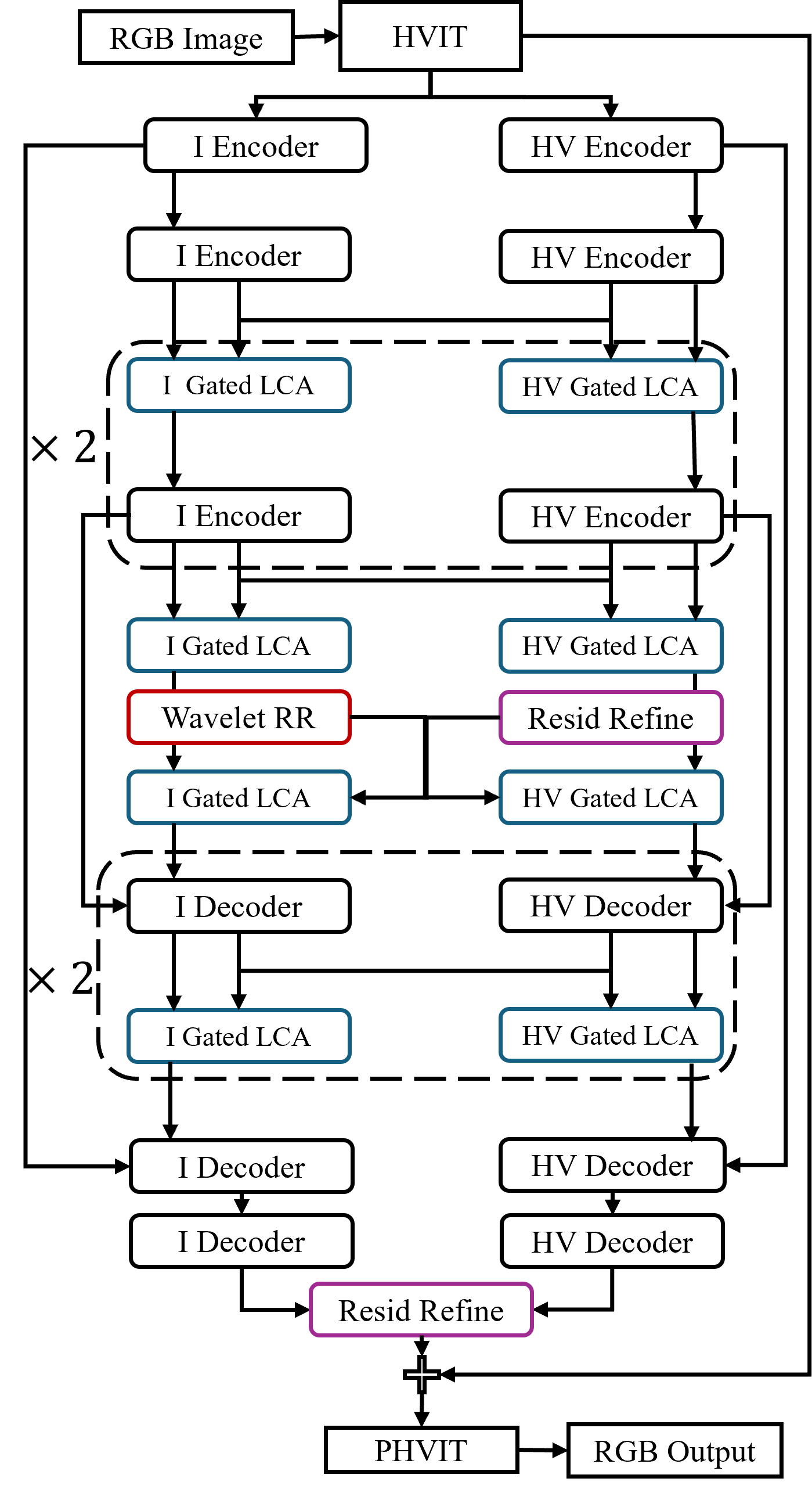}
\caption{Overall structure of WIRNet}
\label{fig:WIRNet}
\end{figure}

\end{document}